\documentclass[letterpaper]{article} % DO NOT CHANGE THIS
\usepackage{aaai25}  % DO NOT CHANGE THIS
\usepackage{times}  % DO NOT CHANGE THIS
\usepackage{helvet}  % DO NOT CHANGE THIS
\usepackage{courier}  % DO NOT CHANGE THIS
\usepackage[hyphens]{url}  % DO NOT CHANGE THIS
\usepackage{graphicx} % DO NOT CHANGE THIS
\usepackage{natbib}  % DO NOT CHANGE THIS AND DO NOT ADD ANY OPTIONS TO IT
\usepackage{caption} % DO NOT CHANGE THIS AND DO NOT ADD ANY OPTIONS TO IT
\usepackage{algorithm}
\usepackage{algorithmic}

\usepackage{newfloat}
\usepackage{listings}
\DeclareCaptionStyle{ruled}{labelfont=normalfont,labelsep=colon,strut=off} % DO NOT CHANGE THIS
\floatstyle{ruled}
\newfloat{listing}{tb}{lst}{}
\floatname{listing}{Listing}
\usepackage{multirow}
\usepackage{bm}
\usepackage{booktabs} 
\usepackage{amsfonts}
\usepackage{amsmath}
\usepackage{amssymb}
\usepackage{mathtools}
\usepackage{amsthm}
\usepackage{nicefrac}
\newcommand{\mac}[1]{{\mathcal #1}}
\newcommand{\mab}[1]{{\mathbb #1}}

\usepackage{stackengine}

\title{Training-Free Time-Series Anomaly Detection: \\ Leveraging Image Foundation Models}
\author {
    Nobuo Namura\textsuperscript{\rm 1},
    Yuma Ichikawa\textsuperscript{\rm 1},
}
\affiliations {
    \textsuperscript{\rm 1} Artificial Intelligence Laboratory, Fujitsu Limited, Japan\\
    namura.nobuo@fujitsu.com, ichikawa.yuma@fujitsu.com
}

\begin{document}

\maketitle

\begin{abstract}
Recent advancements in time-series anomaly detection have relied on deep learning models to handle the diverse behaviors of time-series data. However, these models often suffer from unstable training and require extensive hyperparameter tuning, leading to practical limitations. Although foundation models present a potential solution, their use in time series is limited. To overcome these issues, we propose an innovative image-based, training-free time-series anomaly detection (ITF-TAD) approach. ITF-TAD converts time-series data into images using wavelet transform and compresses them into a single representation, leveraging image foundation models for anomaly detection. This approach achieves high-performance anomaly detection without unstable neural network training or hyperparameter tuning. Furthermore, ITF-TAD identifies anomalies across different frequencies, providing users with a detailed visualization of anomalies and their corresponding frequencies. Comprehensive experiments on five benchmark datasets, including univariate and multivariate time series, demonstrate that ITF-TAD offers a practical and effective solution with performance exceeding or comparable to that of deep models.
\end{abstract}

\section{Introduction}
\label{sec:intro}
Time-series anomaly detection (TAD) is crucial across various fields, including industry, public service, healthcare, and finance, to monitor their status and avoid undesired events such as malfunctions \citep{woike2014structural}, diseases \citep{chauhan2015anomaly, wang2023ecggan}, and frauds \citep{doshi2021timely, zhang2006attack}. 

In practical applications, collecting sufficient abnormal data is often challenging, and the diverse anomaly patterns complicate the use of supervised learning for classifying abnormal and normal data.
Therefore, unsupervised or semi-supervised learning is more suitable, as it can identify anomalies using only unlabeled or normal data and detect previously unseen anomalies without past abnormal data.
% 四種の分類の説明、deep modelが増えていることまでにとどめる
Unsupervised TAD methods can be broadly classified into three categories based on how they evaluate anomaly scores:
(I) Similarity-based methods, which learn the distribution of normal data and use the distance from it or its probability density as the anomaly score \citep{breunig2000lof, scholkopf2001estimating}; (II) Prediction-based methods, which predict the future values of a time series and measure the anomaly score as the discrepancy between predicted and actual values\citep{malhotra2015long};
(III) Reconstruction-based methods, which compress and reconstruct time series and use reconstruction error as the anomaly score \citep{audibert2020usad}. 
While classical approaches detect anomalies based on points in time series, recent studies focus on detecting anomalies through changes or rare patterns in time series. 
Deep learning techniques such as variational autoencoders (VAE) \citep{li2021multivariate}, generative adversarial networks (GAN) \citep{li2019mad}, recurrent neural networks (RNN) \citep{hundman2018smap, shen2020timeseries}, and Transformers \citep{chen2021learning, song2024memto} are frequently employed.
\begin{figure*}[tb]
  \centering
  \includegraphics[width=0.8\textwidth]{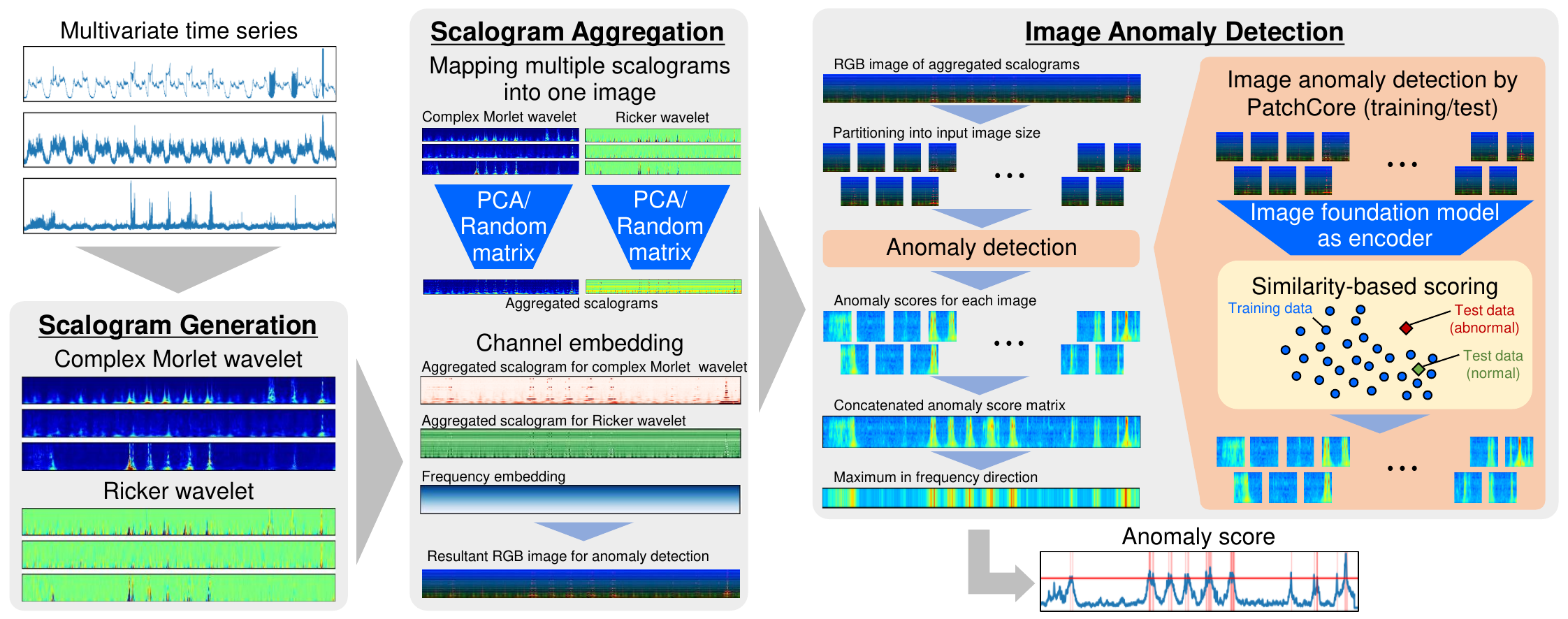}
  \caption{Data processing steps in ITF-TAD}
  \label{fig:method}
\end{figure*}
The use of highly complex deep models for TAD leads to practical issues. 
These models often suffer from unstable training processes, requiring the hyperparameter tuning of a specific model for each case. 
Moreover, These models necessitate a deep understanding of deep learning techniques and significant human costs. 
To tackle this issue, foundation models, trained on large-scale data and applicable to individual problems without additional training, have been employed in natural language processing \citep{touvron2023llama} and computer vision \citep{xiao2023florence}. 
Although recent efforts have adapted foundation models for TAD \citep{das2023decoder, zhou2023one}, achieving high performance still necessitates fine-tuning, which does not completely eliminate training instability and hyperparameter tuning.

To address these issues, we introduce the image-based training-free TAD (ITF-TAD) approach, which applies pre-trained image foundation models for TAD without additional neural network training and fine-tuning. 
ITF-TAD converts time-series data into images using techniques such as continuous wavelet transform (CWT) and then compresses these images into lower-channel representations to enhance computational efficiency. 
These compressed images are subsequently fed into a pretrained image foundation model. 
By aggregating the intermediate layers of the foundation model, a representation space of the time series is obtained, enabling similarity-based anomaly detection on these representations. 
Additionally, this approach identifies the anomaly locations in the frequency space.  
This process is summarized in Fig.~\ref{fig:method}
ITF-TAD demonstrates sufficient performance on five benchmark datasets, proving to be superior or comparable to prior deep models.

\section{Related Works}\label{sec:related-work}
\paragraph{Time-Series Anomaly Detection}
Most unsupervised TAD methods fall into three main categories: similarity-based, prediction-based, or reconstruction-based approaches. 
In similarity-based methods, deep autoencoder on Gaussian mixture latent space (DAGMM) \citep{zong2018deep} stands out.
This method reduces dimensionality for multivariate time series using an autoencoder, then scores anomalies based on probability density.
Other notable methods include support vector data description (SVDD) \citep{tax2004support} and its high-dimensional extension, Deep-SVDD \citep{ruff2018deep}. 
For prediction-based methods, the classical approach is the autoregressive integrated moving average (ARIMA) \citep{Anderson1976TimeSeries2E}. 
Recent studies employ deep models such as long short-term memory (LSTM) \citep{hundman2018smap} and graph deviation network (GDN) \citep{deng2021graph}. 
DiffAD \citep{xiao2023imputation}, which employs a diffusion probabilistic model also falls under this category. Reconstruction-based methods often utilize deep models such as LSTM-VAE \citep{park2018multimodal}, OmniAnomaly \citep{su2019smd}, and Anomaly Transformer \citep{xu2021anomaly}.
GPT2-backbone frozen pre-trained Transformer (GPT2-FPT) \citep{zhou2023one}, which fine-tunes pre-trained language foundation models, is also based on reconstruction.  

\paragraph{Image-Based Approaches}
Traditional frequency analysis methods, such as Fourier and wavelet transforms, have been widely used in practical applications like machine vibration \citep{PENG2004199}, electrocardiograms \citep{khorrami2010comparative}, and electroencephalography \citep{turk2019epilepsy}. 
TimesNet \citep{wu2022timesnet} has employed fast Fourier transform to decompose time series into several characteristic periods, rearranging them in two dimensions for anomaly detection based on reconstruction error. 
This transformation from univariate time series to two-dimensional images is further enhanced using convolutional neural networks (CNN) \citep{aslan2022deep, copiaco2023innovative}, which enhanace explainability by enabling users to infer the causes of anomalies through visualizing abnormal areas or frequencies in images.
Another imaged-based approach demonstrated that scalograms, generated by CWT, outperform other image encoding and a non-encoding TAD methods \citep{garcia2022temporal}. 
The frequency-time characteristics of scalograms also improve human understanding of anomalies.
However, these methods still require extensive and unstable neural network training and hyperparameter tuning. Additionally, while scalogram approaches are effective for univariate time series, extending them to multivariate series presents significant challenges.

\section{Method}
\label{sec:method}
We consider TAD for a $D$-dimensional multivariate time series with length $T$, denoted as $X=(\bm{x}^{d}) \in \mathbb{R}^{D \times T}$, where $\bm{x}^{d} = (x^{d}_{t}) \in \mathbb{R}^{T}$ represents each time-series data in $D$-dimensional multivariate time series.
In TAD, models provide anomaly scores $\bm{\hat{y}} = (\hat{y}_{t}) \in \mathbb{R}^{T}$, which increase when a time point $t$ is likely abnormal. 
As shown in Fig. \ref{fig:method}, the ITF-TAD process includes three main stages: scalogram generation that uses dual mother wavelets, scalogram aggregation achieved by mapping and channel embedding, and anomaly detection using an image foundation model. 
The following sections delve into  each of these three stages.

\subsection{Scalogram Generation with Dual Mother Wavelets}\label{subsec:scalogram}
In CWT, univariate time series of continuous variables are transformed into frequency-time characteristics, known as \textit{scalograms}.This is achieved by translating and scaling a mother wavelet. 
ITF-TAD applies CWT to each dimension $\bm{x}^{d}$ of the time-series data, resulting in $D$ scalograms. 

Consider a moving window size $n$ and pseudo frequencies $\omega$, which are equally distributed on a logarithmic scale, ranging from $\nicefrac{1}{n}$ to $0.5$ Hz and consist of $\Omega$ discrete points. The actual frequency is obtained by multiplying them by the sampling frequency of the time series. CWT can be expressed as follows:
\begin{multline}
    \label{transform}
    S = (S_{i}^{d}) \in \mathbb{R}^{I \times D \times T \times \Omega}, \\
    ~~S_{i}^{d} = (S_{i, t, \omega}^{d}) = \mathrm{CWT}_{\phi_{i}}(\bm{x}^{d}) \in \mathbb{R}^{T \times \Omega}
\end{multline}
where $\mathrm{CWT}_{\phi_{i}}: \mab{R}^{T} \to \mab{R}^{T \times \Omega}$ denote the transformation performed using the aforementioned process with a mother wavelet $\phi_{i}$. 
$S_{i}^{d}$ represents the scalogram of the $d$-th time series. 
This study employs the complex Morlet wavelet \citep{morlet1984cwt} and the Ricker wavelet \citep{ricker1953form} as $\phi_{i}$. 
% The scalograms for the complex Morlet wavelet are computed as the absolute values of real and imaginary parts.
The dual use of complex and real mother wavelets is essential for maintaining high sensitivity in both frequency characteristics and phase shifts.
Scalograms generated with complex mother wavelets exhibit high sensitivity in frequency but lose temporal accuracy while those with real wavelets, such as the Ricker wavelet, exhibit high sensitivity to phase shifts and lower frequency sensitivity. 
Additionally, these scalograms are normalized by using the maximum absolute value of each scalogram as follows:
\begin{equation}
\label{eq:normalize}
    \bar{S}^{d}_{i, t, \omega} = \frac{S^{d}_{i, t, \omega}}{\displaystyle \max_{t, \omega} |S^{d}_{i, t, \omega}|}
\end{equation}
This normalization keeping the origin at zero is essential for leveraging the sparsity of scalograms during aggregation. 

\subsection{Scalogram Aggregation}
\label{subsec:aggregation}
To use image foundation models for TAD, the normalized scalograms $\bar{S} \in \mab{R}^{I \times D \times T \times \Omega}$ need to be converted into the input format $\hat{S} \in \mab{R}^{\hat{I} \times \hat{T} \times \hat{\Omega}}$ using a mapping $\mac{F}: \mab{R}^{I \times D \times T \times \Omega} \to \mab{R}^{\hat{I} \times \hat{T} \times \hat{\Omega}}$ to perform TAD.
Typically, image foundation models require, $\hat{I} \leq 3$.
It is important to maintain the time dimension $\hat{T}=T$ during this conversion to accurately identify anomalous segments.
This mapping process remains essential even when using other foundation models, such as language foundation models. 
In the following, we briefly introduce the mapping function $\mac{F}$ for pre-trained image foundation models. 

\paragraph{Component-Wise PCA Mapping}
When using pre-trained image foundation models, it is crucial to ensure that the condition $\hat{I} \le 3$ and $\hat{T} = T$ holds true, resulting in $I \times D \times T \times \Omega > \hat{I} \times \hat{T} \times \hat{\Omega}$. 
This necessitates information compression.
To efficiently minimize information loss during compression, we employ a principal component analysis (PCA) for each $T$ component, refered to as \textit{component-wise PCA mapping}. 
Specifically, for any $i = 1, \ldots, I$,  $\mac{F}_{\mathrm{PCA}_{i}}: \mathbb{R}^{I \times D \times T \times \Omega} \to \mathbb{R}^{I \times T \times \hat{\Omega}}$, where the matrix $W_{i}^\mathrm{PCA} \in \mathbb{R}^{(D \times \Omega) \times \hat{\Omega}}$ denotes the concatenated eigenvectors corresponding to the first $\hat{\Omega}$ principal  components. 
This PCA mapping considers $D \times \Omega$ dimensional feature vector and $T$ as the number of data points for 
each component $i=1, \ldots, I$.
This component-wise PCA mapping ensures consistency in the time direction and alignment with the mother wavelet before and after transformation
The aggregated scalogram $S_{i}^{\mathrm{agg}}$ is given by 
\begin{equation}
\label{eq:pca}
    S_{i}^{\mathrm{agg}} 
    = \mac{F}_{\mathrm{PCA}_{i}}\left(\mathrm{Concat}\left[\{\bar{S}^{d}_{i}\}_{d=1}^{D} \right] \right) 
    \in \mathbb{R}^{T \times \hat{\Omega}}. 
\end{equation} 
where $\mathrm{Concat}$ represents the operation of concatenating $\bar{S}^{d}_{i}$ along the frequency $\omega$ direction.
The frequency resolution $\Omega$ of the CWT is set so that $\Omega \leq \nicefrac{T}{D}$. 
Although learnable approaches such as autoencoders could achieve this, we avoid the neural network training and the hyperparameter tuning by using PCA. 

\paragraph{Random Mapping}
In TAD, the backbone structure of time-series data, characterized by principal components with high contribution rates, does not necessarily indicate anomalies. While this structure is common, anomalies are often characterized by slight deviations from this backbone. Component-wise PCA mapping emphasizes components with high contribution rates, highlighting variations in the primary structure but may overlook minor anomalies outside this structure. 
To address this limitation, we propose another mapping method, refered to as \textit{random mapping}. Here, the aggregated scalograms $S_{i}^{\mathrm{agg}}$ are obtained by the following dimensionality reduction using linear transformation with a random tensor $\Gamma =(\gamma_{i,d,\omega}) \in \mab{R}^{I \times D \times \Omega}$:
\begin{equation}
\label{eq:aggregate}
    S^{\mathrm{agg}} \in \mab{R}^{I\times T \times \Omega},~~
    S_{i,t,\omega}^{\mathrm{agg}} = \frac{\sum_{d=1}^{D} \gamma_{i,d,\omega} \bar{S}_{i,t,\omega}^{d}}{\sum_{d=1}^{D} \gamma_{i,d,\omega}}. 
\end{equation}
Specifically, for each $i$-th mother wavelet, this random tensor is uniformly generated from the Latin hyper cube, characterized by less-than-or-equal-to $\max(2, \lfloor \nicefrac{D}{p} \rfloor)$ elements to be $1$, from the $\omega$-th column of $\gamma_{i, d, \omega}$.
We set $p=5$ in this study. Detailed ablation studies of this parameter p are provided in the Appendix \ref{apx:rand}. 
However, this random mapping assumes that scalograms are typically sparse. If they are not, the aggregated results become unsuitable for effective TAD.

\paragraph{Channel Embedding}
Next, the aggregated scalogram $S_{i}^{\mathrm{agg}}$ is transformed into a format compatible with pretrained image foundation models,
specifically RGB images, i.e, $\hat{I} = 3$. 
As illustrated in Fig. \ref{fig:method}, two channels (Red, Green) are used to input $S_{i}^{\mathrm{agg}}$, while the remaining channel (Blue) inputs frequency indexes $\hat{\omega}$, referred to as \textit{frequency encoding} in this study. 
This frequency encoding functions similarly to the positional encoding \citep{vaswani2017attention} employed in transformers. 
This design is crucial because most image anomaly detection models decompose each image into numerous patches and perform anomaly detection, often ignoring positional information related to frequency, which is crucial for anomaly segmentation.
As such models cannot differentiate where learned normal image patches are located within a test image, false positives can occur if similar patches are located in different frequency bands. 
By incorporating frequency information in the image, any image anomaly detection model can more accurately detect such anomalies. 

Additionally, the aggregated scalograms is normalized for imaging as
\begin{align}
\label{eq:imaging}
    \bar{S}_{i,t,\omega}^{\mathrm{agg}} &= \max \left[0, \min \left(1, 
    \frac{S_{i,t,\omega}^{\mathrm{agg}}/r - S_{i}^{\min}}{S_{i}^{\max} - S_{i}^{\min}} 
    \right) \right] \notag,\\~
    S_{i}^{\min} &= \min \left[0, \min_{t,\omega} S_{i,t,\omega}^{\mathrm{agg}} \right],~~
    S_{i}^{\max} = \max_{t,\omega} S_{i,t,\omega}^{\mathrm{agg}} \notag, 
\end{align}
where $r=1.2$ is introduced to provide additional space for test data with larger absolute values in scalograms, and to identify them as anomalies. 
Finally, the values in each channel are converted into $\{0, 1, \cdots, 255\}$ for imaging.

\subsection{Anomaly Detection with Image Foundation Model}
By converting time-series data into images, any image anomaly detection model can be applied to TAD.
In this study, we use ITF-TAD with an image foundation model, specifically employing PatchCore \citep{roth2022towards} for image anomaly detection. PatchCore is an semi-supervised learning model that detects anomalies in images without needing to train a new neural network. It utilizes a pretrained ResNet\citep{he2016deep}-type classifier as a foundation model to extract feature vectors from images. During the training process, the model selects representative points, known as a \texttt{coreset}, from the feature vectors of the training data. Anomaly scores are then calculated based on the distances from the nearest points in the coreset. 

\paragraph{Image Dividing}
In PatchCore, image features are extracted using a ResNet-type classifier trained on ImageNet \citep{deng2009imagenet}. 
This means the achievable temporal resolution for anomaly detection matches the resolution of the classifier's input images. In this study, we use \texttt{PatchCore} as implemented in Anomalib\citep{akcay2022anomalib} employing a pretrained Wide ResNet-50-2, whose resolution is 256, provided by Torchvision. Therefore, the entire image must be divided using a window size of $n\leq256$. We chose $n=256$ and set the number of discrete points in the frequency direction to $\Omega=min(n, \nicefrac{T}{D})$ and $\hat{\Omega}=n$ for PCA mapping. $\Omega=\hat{\Omega}=n$ is used for random mapping. While previous studies \citep{wu2022timesnet, xu2021anomaly} used non-overlapping windows, our preliminary tests with PatchCore for ITF-TAD showed missed anomalies at image edges.  To address this, we set the stride width to $s=\nicefrac{n}{2}$. 
% Effects of $n$ and $m$ on performances are investigated in Appendix \ref{apx:window}.

\paragraph{Anomaly Score Computation}
After creating images for both training and test data using the described process, PatchCore evaluates the anomaly score for each pixel and converts these scores back into a time series. For simplicity, we first consider the case where the time series is divided without overlap ($s=n$). Here the anomaly score for each image forms an $n \times \hat{\Omega}$ matrix. When $s=n$, concatenating these anomaly score matrices in the time direction yields a large anomaly score matrix $A_{t,\hat{\omega}}$. At each time index $t$, the anomaly score is defined using the maximum value in the frequency direction as $\hat{y}_t = \max_{\hat{\omega}} A_{t,\hat{\omega}}$. If multiple images share the same $t$ ($s<n$), the highest anomaly score among them is taken as the final anomaly score.. 

When using scalograms, corrections are needed at the start and end points of the anomaly scores in the test data. Near the temporal edges of the scalogram, accurate expression of frequency-time characteristics is challenging owing to the influence of the time series' end, known as the cone of interference. In practical applications, these edge regions are typically excluded from anomaly detection evaluation. Therefore, in this study, after calculating the anomaly scores, the scores within $n$ points from the start and end of the test data are replaced with the minimum anomaly score found within the training and test data. This method is replicates the scenario where anomalies cannot be detected in these regions.

For actual anomaly detection, arbitrary threshold $\delta$ is used to determine anomalies, where time indexes with $\hat{y}_t \geq \delta$ are considered anomalous. The decision of this threshold depends on the availability of real data specific to each case, especially the availability of anomaly data, and thus is not discussed here. In the benchmark datasets used in this study, the threshold that maximizes the F1 score is adopted, as seen in previous research..

\section{Experiments}

\subsection{Experimental Setup}
\label{subsec:setup}

\paragraph{Benchmark Datasets}
To evaluate ITF-TAD performance, we use five types of univariate and multivariate time-series benchmark datasets including UCR \citep{keogh2021ucr}, PSM \citep{abdulaal2021psm}, SMAP \citep{hundman2018smap}, MSL \citep{hundman2018smap}, SMD \citep{su2019smd}. Details of these datasets are summarized in Table \ref{tab:dataset}. Missing value sections included in the training interval of the PSM dataset are excluded. A duplicated subdataset in SMAP is omitted. Additional features of each dataset are provided in Appendix \ref{apx:dataset}.

\begin{table*}[tb]
\caption{Dataset statistics.}
\label{tab:dataset}
\centering
\begin{tabular}{c|ccccc}
\toprule
Dataset & Dimensions & Subdatasets & \#Training data & \#Test data & Anomaly rate {[}\%{]} \\
\midrule
UCR     & 1          & 250      & 5302449  & 14051317 & 0.35               \\
PSM     & 25         & 1        & 129784   & 87841    & 27.76              \\
SMAP    & 25         & 54       & 138004   & 435826   & 12.82              \\
MSL     & 55         & 27       & 58317    & 73729    & 10.48              \\
SMD     & 38         & 28       & 708405   & 708420   & 4.16              \\
\bottomrule
\end{tabular}
\end{table*}

\paragraph{Baselines}
For comparison with ITF-TAD, we selected seven state-of-the-art baselines: GPT2-LFT \cite{zhou2023one}, TimesNet \cite{wu2022timesnet}, DiffAD \cite{xiao2023imputation}, TranAD \cite{tuli2022tranad}, Anomaly Transformer \cite{xu2021anomaly}, GDN \cite{deng2021graph}, and LSTM \cite{hundman2018smap}. All these baselines are based on deep learning and require neural network training or fine-tuning. If the hyperparameters for each dataset were specified in the author's source code, those values were used. For datasets not covered in the papers, hyperparameters from similar datasets were applied, referencing the subdataset's data size. Detailed information on these settings is provided in Appendix \ref{apx:hyperparam}.

\paragraph{Hyperparameters}
The hyperparameter values for ITF-TAD consistent across all datasets. Additionally, the parameters used within PatchCore in ITF-TAD are consistent across all. Specifically, the the coreset sampling ratio is set to 0.01, and the number of neighbors used for calculating anomaly scores is set to 9.

\paragraph{Preprocessing of Anomaly Score before Evaluating Metrics}

\begin{figure}[tb] 
  \centering
  \includegraphics[width=0.45\textwidth]{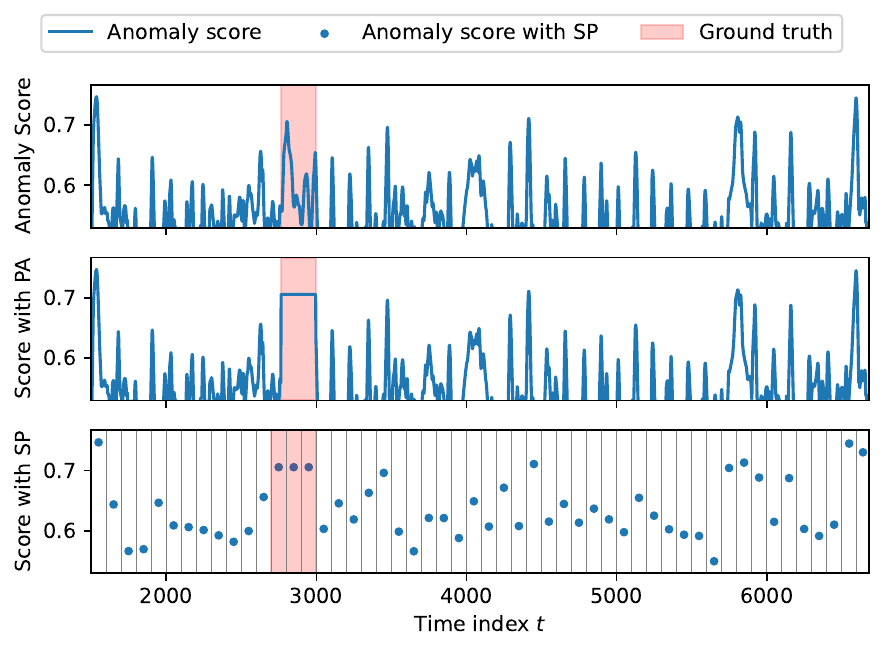}
  \caption{Score partitioning on UCR-053}
  \label{fig:sp}
\end{figure}

To ensure fair comparison using common metrics such as F1 score and AUCPR, we introduce a preprocessing method called score partitioning (SP) for anomaly scores. In prior research, a correction known as point adjustment (PA) has been frequently applied when calculating evaluation metrics. PA considers an entire segment of consecutive anomalies as identified if at least one point within the segment is recognized as anomalies. PA itself is a practical and reasonable correction in existing benchmark datasets, where anomaly labels are excessively lengthy or encompass normal intervals, leading to a high occurrence of false negatives without PA (see Appendix \ref{apx:vis} for details). However, as pointed out in \cite{doshi2022reward, Kim2022}, using PA alone can be flawed, as it may achieve state-of-the-art performance even with random anomaly scores. To address flaw, we combine SP with PA to ensure a fair comparison.
The process and effects of SP on anomaly scores are illustrated in Fig. \ref{fig:sp}.
First, we apply PA to anomaly scores by substituting each interval labeled as abnormal in the ground truth with the maximum anomaly score found in that interval. Following this, we employ SP to divide the anomaly scores into non-overlapping sections with a window size of $n_{sp}=100$. The highest anomaly score within each section is then assigned as the representative score for that section. For ground truth labels, any section containing at least one abnormal point is classified as abnormal. Since PA is applied in advance, segmenting an abnormal interval does not result in an increase in false negatives, even if labels are improperly assigned. We calculate metrics based on these representative anomaly scores.

\paragraph{Evaluation Metrics}
For evaluation metrics, we employ the best F1 score and AUCPR with SP, denoted as F1*-SP and AUCPR-SP. F1*-SP represents the highest F1 score achieved with SP when the threshold varies, and this optimal threshold is used for visualizing anomaly scores. Both F1*-SP and AUCPR-SP are calculated for each subdataset, with their averages used for comparison across datasets. In the UCR dataset, which contains a single artificially added anomaly in each subdataset, the number of correct answers serves as an evaluation metric. This metric counts the subdatasets where the highest anomaly score falls within a $\pm100$ range of the ground truth segment, a method used for ranking participants in the KDD Cup 2021. Since these evaluation metrics alone may not fully capture model performance, we also visualize representative cases through time series of anomaly scores for a more detailed comparison.

\subsection{Main Results}
The results of applying ITF-TAD and seven baselines to each benchmark dataset are listed in Table \ref{tab:results}. ITF-TAD comes in two variants: ITF-TAD-PCA, which uses PCA, and ITF-TAD-RM, which uses random matrices for mapping. For univariate time series like the UCR dataset, only ITF-TAD-PCA metrics are documented. ITF-TAD-RM, using random matrices, was run five times with different random seeds, and the mean metrics are reported. Table \ref{tab:results} also includes results from the two-sided Wilcoxon signed rank test without multiple testing correction for datasets consisting of multiple subdatasets (UCR, SMAP, MSL, and SMD) as five types of symbols. The symbols ``$\gg$''$/$``$>$'' and ``$\ll$''$/$``$<$'' indicate ITF-TAD-PCA has significantly higher and lower results at significance levels of $0.01/0.05$, while the symbol ``$\approx$'' indicates there is no significant difference. Metrics other than those in Table \ref{tab:results} are summarized in Appendix \ref{apx:metrics}.

\begin{table*}[tb]
\caption{Performance comparison among ITF-TAD and baselines on five benchmark datasets. The best results are underlined, while the top two results are in bold. }
\label{tab:results}
\centering
\scalebox{0.75}{
\begin{tabular}{cc|ccccccccc}

\toprule
\multirow{2}{*}{Dataset} & \multirow{2}{*}{Metric} & \multicolumn{2}{c}{ITF-TAD (Ours)}        & GPT2-FPT             & TimesNet            & DiffAD & TranAD               & A.Trans. & GDN                  & LSTM  \\
                         &                         & PCA                  & RM                   & (2023)               & (2023)              & (2023) & (2022)               & (2022)   & (2021)               & (2018) \\
\midrule
\multirow{3}{*}{UCR}     & F1*-SP                  & \multicolumn{2}{c}{\underline{ \textbf{0.724}}}    & $\gg$ 0.231                & $\gg$ 0.244               & $\gg$ 0.199  & $\gg$ 0.306                & $\gg$ 0.237    & $\gg$ \textbf{0.512}       & $\gg$ 0.197 \\
                         & AUCPR-SP                & \multicolumn{2}{c}{\underline{ \textbf{0.819}}}    & $\gg$ 0.504                & $\gg$ 0.521               & $\gg$ 0.487  & $\gg$ 0.560                 &$\gg$ 0.526    & $\gg$ \textbf{0.671}       & $\gg$ 0.513 \\
                         & Correct                 & \multicolumn{2}{c}{\underline{ \textbf{150}}}      & 22                   & 33                  & 23     & 37                   & 17       & \textbf{99}          & 20    \\
\midrule
\multirow{2}{*}{PSM}     & F1*-SP                  & \underline{ \textbf{0.889}} & \underline{ \textbf{0.889}} & 0.886                & 0.860                & 0.821  & 0.851                & 0.844    & 0.812                & 0.569 \\
                         & AUCPR-SP                & 0.915                & 0.895                & \underline{ \textbf{0.949}} & \textbf{0.934}      & 0.842  & 0.901                & 0.913    & 0.883                & 0.435 \\
\midrule
\multirow{2}{*}{SMAP}    & F1*-SP                  & 0.720                & $\approx$ 0.730                & $>$ 0.651                & $>$ 0.628               & $\gg$ 0.518  & $\approx$ \textbf{0.731}       & $\gg$ 0.577    & $\approx$ \underline{ \textbf{0.750}}  & $\approx$ 0.620  \\
                         & AUCPR-SP                & 0.797                & $\approx$ \textbf{0.798}       & $\gg$ 0.719                & $\gg$ 0.715               & $\gg$ 0.642  & $\approx$ \underline{ \textbf{0.803}} & $\gg$ 0.681    & $\approx$ \textbf{0.798}       & $>$ 0.685 \\
\midrule
\multirow{2}{*}{MSL}     & F1*-SP                  & \underline{ \textbf{0.727}} & $\approx$ 0.724                & $\approx$ 0.606                & $\approx$ 0.659               & $\approx$ 0.598  & $\approx$ 0.713                & $>$ 0.575    & $\approx$ \underline{ \textbf{0.727}} & $\approx$ 0.719 \\
                         & AUCPR-SP                & 0.737                & $\approx$ \textbf{0.756}       & $>$ 0.637                & $\gg$ 0.672               & $\approx$ 0.626  & $\approx$ 0.751                & $>$ 0.616    & $\approx$ \underline{ \textbf{0.768}} & $\approx$ 0.727 \\
\midrule
\multirow{2}{*}{SMD}     & F1*-SP                  & 0.645                & $\approx$ 0.660                & $\ll$ \textbf{0.844}       & $\ll$ \underline{ \textbf{0.850}} & $\gg$ 0.330   & $\ll$ 0.800                  & $\approx$ 0.602    & $\ll$ 0.808                & $\approx$ 0.571 \\
                         & AUCPR-SP                & 0.585                & $\approx$ 0.620                & $\ll$ \underline{ \textbf{0.833}} & $\ll$ \textbf{0.829}      & $\gg$ 0.266  & $\ll$ 0.790                 & $\approx$ 0.542    & $\ll$ 0.779                & $\approx$ 0.514 \\
\bottomrule
\end{tabular}
}
\end{table*}

\begin{table*}[tb]
\caption{Ablation study for channel embedding on five benchmark datasets. The best results appear in bold. ``(R/G/B)'' means the channels used.}
\label{tab:ablation}
\centering
\scalebox{0.75}{
\begin{tabular}{cc|cccccc}
\toprule
\multirow{2}{*}{Dataset} & \multirow{2}{*}{Metric} & \multicolumn{6}{c}{ITF-TAD-PCA}                                                                                           \\
                         &                         & Full           & w/o Morlet         & w/o Ricker              & w/o Frequency  & only Ricker             & only Morlet    \\
                         &                         & (RGB)            & (GB)                 & (RB)                      & (RG)             & (G)                       & (R)              \\
\midrule
\multirow{3}{*}{UCR}     & F1*-SP                  & 0.724          & $\approx$ 0.732    & $\approx$ 0.727          & $<$ 0.746      &       $<$ 0.743      & \textbf{$\ll$ 0.777}    \\
                         & AUCPR-SP                & 0.819          & $\approx$ 0.828     & $\approx$ 0.808          & $<$ 0.835      & $\ll$ 0.838    & \textbf{$<$ 0.847}      \\
                         & Correct                 & 150            & 149                & 151                     & 152            & 151                     & \textbf{163}   \\
\midrule
\multirow{2}{*}{PSM}     & F1*-SP                  & \textbf{0.889} & 0.873              & 0.833                   & 0.863          & 0.882                   & 0.836          \\
                         & AUCPR-SP                & 0.915          & 0.923              & 0.860                   & 0.903          & \textbf{0.924}          & 0.911          \\
\midrule
\multirow{2}{*}{SMAP}    & F1*-SP                  & \textbf{0.720} & $>$ 0.607          & $\gg$ 0.542             & $\gg$ 0.587    & $\gg$ 0.598             & $\gg$ 0.588    \\
                         & AUCPR-SP                & \textbf{0.797} & $>$ 0.718          & $\gg$ 0.657             & $\gg$ 0.691    & $\gg$ 0.719             & $\gg$ 0.680    \\
\midrule
\multirow{2}{*}{MSL}     & F1*-SP                  & 0.727          & $\approx$ 0.715     & $\approx$ 0.730          & $\approx$ 0.687 & \textbf{$\approx$ 0.744} & $\approx$ 0.671 \\
                         & AUCPR-SP                & 0.737          & $\approx$ 0.744     & \textbf{$\approx$ 0.770} & $\approx$ 0.705 & $\approx$ 0.744          & $\approx$ 0.707 \\
\midrule
\multirow{2}{*}{SMD}     & F1*-SP                  & 0.645          & \textbf{$<$ 0.683} & $\approx$ 0.645          & $>$ 0.620      & $\approx$ 0.657          & $\approx$ 0.642 \\
                         & AUCPR-SP                & 0.585          & \textbf{$<$ 0.640} & $\approx$ 0.563          & $>$ 0.560      & $\approx$ 0.595          & $\approx$ 0.574 \\
\bottomrule
\end{tabular}
}
\end{table*}

ITF-TAD achieves significantly higher performance in the UCR dataset and comparable results with the best baseline models in PSM, SMAP, and MSL, all without requiring any neural network training. Specifically, the number of correct answers in UCR is remarkable; its correct answer rate of 0.6 (150/250) also outperforms the best-reported value of 0.47 by \citet{rewicki2023worth}. This result indicates that scalograms are more informative than raw time-series data themselves and that combining scalograms with an image foundation model yields high-performance TAD.
The differences between ITF-TAD-PCA and ITF-TAD-RM are minor. PCA is more computationally intensive for eigenvalue decomposition. Thus, ITF-TAD-RM should be more reasonable for real-world applications. The computational times for each model are shown in Appendix \ref{apx:time}.

\begin{figure*}[tb]
  \centering
  \includegraphics[width=0.8\linewidth]{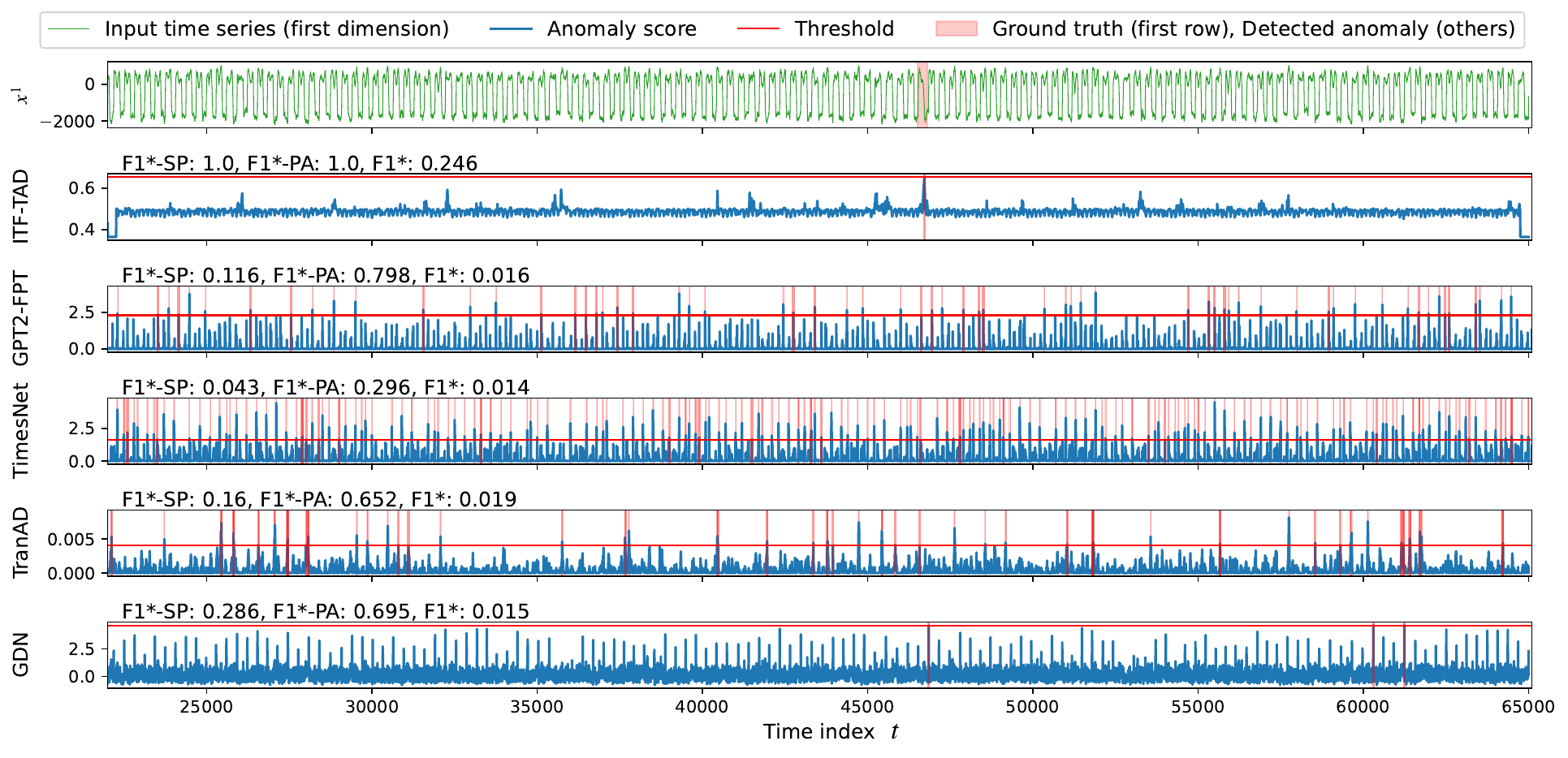}
  \caption{Anomaly scores from five models for the test section of UCR-060}
  \label{fig:ucr}
\end{figure*}

\begin{figure*}[tb]
  \centering
  \includegraphics[width=0.8\linewidth]{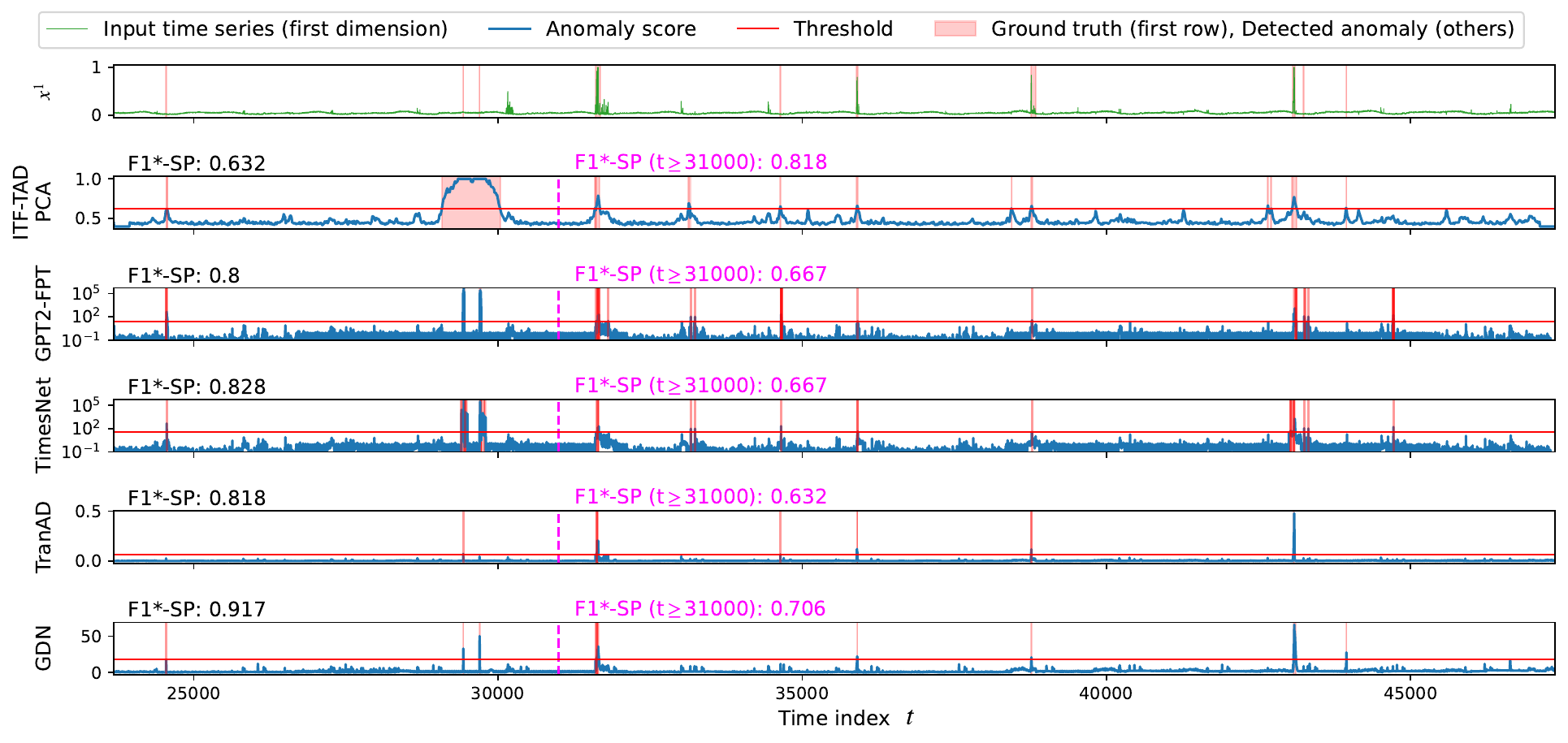}
  \caption{Anomaly scores from five models for the test section of SMD-2-3}
  \label{fig:smd}
\end{figure*}
Figure \ref{fig:ucr} presents a case from the UCR dataset (UCR-060), showing the time series $\bm{x}^{1}$ and anomaly scores from ITF-TAD, GPT2-FPT, TimesNet, TranAD, and GDN for the test section. The threshold used maximizes the F1*-SP shown in the top left of each model's anomaly scores. Only ITF-TAD's anomaly scores clearly spike at the anomaly labels. On the other hand, GDN generates two segments of false positives, and the other models generate numerous false positives. They have sufficiently low F1*-SP as intended, while the best F1 scores with PA alone (F1*-PA) shown in Fig. \ref{fig:ucr} are improperly high. These results demonstrate the validity of the correction using SP.

ITF-TAD has modest results in the SMD dataset, primarily owing to the unique distributions of its anomaly scores, which cause underestimations of its performance. SMD is characterized by the presence of numerous relatively short anomaly labels within each subdataset. Fig. \ref{fig:smd} shows the time series $\bm{x}^{1}$ and anomaly scores from ITF-TAD-PCA, GPT2-FPT, TimesNet, TranAD, and GDN for the test section of SMD-2-3. Although all methods detect many anomalies with few false detections, ITF-TAD-PCA has a particularly low F1*-SP score. This is because scalograms, which ITF-TAD uses, produce has a wider distribution of anomaly scores around peaks, such as those near $t=30000$, increasing the number of false positives in metric calculations. As shown in magenta text in Fig. \ref{fig:smd}, ITF-TAD-PCA achieves the highest F1*-SP score in the interval $t \geq 31000$, suggesting that the wider distribution of anomaly scores can understimate ITF-TAD's performance. Similar score distributions are observed in 22 out of 28 subdatasets of SMD. Although using SP can alleviate this underestimation, excessively increasing the window size $n_{sp}$ for the SP metrics makes accurate evaluation difficult. Hence, qualitative comparisons with visualization are needed. The effects of $n_{sp}$ on performance comparison are summarized in Appendix \ref{apx:sp}.

\subsection{Ablation Study}
To validate the effects of dual aggregated scalograms and frequency embedding in the image via channel embedding, we conducted anomaly detection by removing each channel from ITF-TAD-PCA. The results, shown in Table \ref{tab:ablation}, confirm the usefulness of each channel through the F1*-SP and AUCPR-SP metrics for SMAP, and the F1*-SP for PSM. Notably, removing the aggregated scalogram of the Ricker wavelet largely degrades performance, suggesting it is the most effective component for image-based TAD. 

In the univariate UCR dataset, simpler images, such as those using only the complex Morlet wavelet, have shown improved performance. This suggests that in univariate time series, the image needs less information compared to multivariate cases, making simplicity and clarity more important. This principle may also enhance performance in multivariate TAD by focusing on clearer image mapping.
Additionally, in the SMD dataset, performance improves when the aggregated scalogram of the complex Morlet wavelet is removed. Scalograms of the complex Morlet wavelet often have a wide distribution in the time direction in the lower frequency region, at the bottom of the scalograms, as shown in Fig. \ref{fig:method}. This contributes to the wider distribution of anomaly scores observed in Fig. \ref{fig:smd}. Therefore, removing it led to improved performance. Although the scalogram of the Ricker wavelet also has a wide distribution, the issue of underestimation still persists.

\section{Conclusions}
\label{sec:conclusion}
This study introduces the ITF-TAD approach, which leverages image foundation models to efficiently detect anomalies in time-series data without requiring neural network training. ITF-TAD transforms each dimension of a time series into scalograms using CWT with two types of mother wavelets. These scalograms are then aggregated into a single aggregated scalogram through PCA or random matrices. Using PatchCore as the image anomaly detection model and a pretrained ResNet-type image classifier as the foundation model, ITF-TAD achieves superior or comparable performance across five univariate and multivariate time-series benchmark datasets all without the need for neural network training. 
However, the study acknowledges some limitations. The methods for generating aggregated scalograms without neural network training still have room for improvement. Additionally, the computation time for CWT and anomaly detection using PatchCore is not significantly different from existing deep models, indicating that the full benefits of using foundation models have yet to be realized. Future work aims to establish criteria for information compression suitable for TAD to introduce more effective mapping methods. We also plan to explore the use of foundation models from other fields to develop more time-efficient training-free TAD solutions.


\bibliography{ref}

\clearpage
\appendix
\section*{Appendices}
\section{Dataset Details}
\label{apx:dataset}
\paragraph{UCR}
The UCR dataset\citep{keogh2021ucr}, adopted for the KDD Cup 2021, represents a collection of univariate time series comprising 250 subdatasets. It consists of time-series data from various fields and contains a single artificially added anomaly in each subdataset. Despite being univariate, the clarity and reliability of its anomaly labels owing to artificial anomalies make it suitable for assessing the compatibility of the image-based TAD. 

\paragraph{PSM}
The pooled server metrics (PSM) dataset \citep{abdulaal2021psm} is collected internally from multiple application server nodes at eBay and includes 25 dimensions with no subdatasets. Anomalies appear in both the training and test sections, but labels are only available for the test section. This set-up requires performance evaluation as an unsupervised method rather than a semi-supervised method.

\paragraph{SMAP \& MSL}
The soil moisture active passive (SMAP) satellite and Mars science laboratory (MSL) rover datasets\citep{hundman2018smap}, collected by NASA, are related to spacecraft and contain 25 and 55 dimensions, respectively. The SMAP dataset has 54 subdatasets (though it is to be 55, one is omitted due to duplication) while the MSL dataset includes 27 subdatasets. In both datasets, the first dimension is usually continuous, while the remaining dimensions are often binary, with some subdatasets containing discrete values even in the first dimension. Although CWT may not be suitable for subdatasets with many binary values, these spacecraft-related datasets are rare and widely used. 

\paragraph{SMD}
The server machine dataset (SMD) \citep{su2019smd} is a 38-dimensional dataset acquired by a telecommunications operator and contains 28 subdatasets. This dataset was chosen because the proportion of anomalies it contains is relatively low, usually allowing for reasonable comparisons. 

% 必須
\section{Hyperparameters Used in Baselines}
\label{apx:hyperparam}
The hyperparameters for the baselines are detailed in Table \ref{tab:hyp}. All source codes were downloaded from the authors' GitHub repositories. When configuration files for specific datasets were available, those settings were used. For 
TranAD and GDN, the default parameters were applied across datasets owing to the absence of dataset-specific configuration files. However, for the univariate UCR dataset, the topk parameter for GDN was set to 1. 
For many baselines lacking UCR-specific configurations, settings from the SMD dataset, which has similar time-series lengths, were used. Parameters dependent on the number of dimensions $D$ were adjusted to 1 for UCR. For TimesNet, the number of epochs was set to 3 for UCR  owing to computational constraints. 
The Anomaly Transformer used a common configuration for PSM, SMAP, and MSL, which was also UCR. 
For LSTM, most hyperparameters remained at their default values, but epochs and lstm\_batch\_size were specified in Table \ref{tab:hyp} to account for computational time. For UCR, the values in the table were used according to the length of each subdataset.

\begin{table*}[ht]
\caption{Hyperparameter setting for baselines.}
\label{tab:hyp}
\centering
\scalebox{0.8}{
\begin{tabular}{cl|lllll}
\toprule
Dataset                   &                      & UCR                     & PSM      & SMAP         & MSL          & SMD      \\
\midrule
GPT2-FPT                  &                      & same as SMD             & provided & provided     & provided     & provided \\
\midrule
\multirow{2}{*}{TimesNet} &                      & train\_epochs: 3         & provided & provided     & provided     & provided \\
                          &                      & others: same as SMD     &          &              &              &          \\
\midrule
DiffAD                    &                      & same as SMD             & provided & provided     & provided     & provided \\
\midrule
TranAD                    &                      & default                 & default  & default      & default      & default  \\
\midrule
A.Trans.                  &                      & same as PSM, SMAP, MSL  & provided & provided     & provided     & provided \\
\midrule
GDN      &  & topk: 1, othres: default & default  & default      & default      & default  \\
\midrule
\multirow{3}{*}{LSTM}     & epochs               & 4 or 10                 & 4        & 20           & 20           & 20       \\
                          & lstm\_batch\_size      & 128, 512, 1024, or 2048 & 2048     & 64 (default) & 64 (default) & 512      \\
                          & others               & default                 & default  & default      & default      & default \\
\bottomrule
\end{tabular}
}
\end{table*}

% 必須
\section{Computational Resources and Runtime}
\label{apx:time}
The numerical experiments were conducted on two types of server machines, as detailed in Table 5, which also outlines the benchmark datasets used.

\begin{table*}[ht]
\caption{Computational resources used.}
\label{tab:comp}
\centering
\scalebox{0.75}{
\begin{tabular}{c|cc}
\toprule
Node name       & V100                                             & A100                                                  \\
\midrule
GPU             & NVIDIA V100 for NVLink 16GiB HBM2                & NVIDIA A100 for NVLink 40GiB HBM2                     \\
CPU             & Intel Xeon Gold 6148 Processor 2.4 GHz, 20 Cores & Intel Xeon Platinum 8360Y Processor 2.4 GHz, 36 Cores \\
Memory [GiB]    & 32                                               & 32                                                    \\
Dataset applied & UCR, SMAP, MSL, SMD                              & UCR-239 to 241, PSM        \\
\bottomrule
\end{tabular}
}
% \end{table*}
\vspace{4mm}
% \begin{table*}[ht]
\caption{Runtime for three datasets in minutes.}
\label{tab:time}
\centering
\scalebox{0.75}{
\begin{tabular}{cc|ccccccccc}
\toprule
\multirow{2}{*}{Dataset} & \multirow{2}{*}{Part} & \multicolumn{2}{c}{ITF-TAD (Ours)} & GPT2-FPT & TimesNet & DiffAD & TranAD & A.Trans. & GDN    & LSTM   \\
                         &                       & PCA              & RM              & (2023)   & (2023)   & (2023) & (2022) & (2022)   & (2021) & (2018) \\
\midrule
\multirow{2}{*}{PSM}     & Entire                & 16.2             & 14.2            & 62.6     & 58.7     & 35.1   & 2.8    & 9.6      & 8.8    & 15.8   \\
                         & Imaging                   & 10.7             & 8.4             & -         & -         & -       & -       & -         & -       & -       \\
\midrule
\multirow{2}{*}{SMAP}    & Entire                & 54               & 58.6            & 224.6    & 227.2    & 99.5   & 13.3   & 26.7     & 23.4   & 154.3  \\
                         & Imaging                   & 24.2             & 26.5            & -         & -         & -       & -       & -         & -       & -       \\
\midrule
\multirow{2}{*}{SMD}     & Entire                & 574.8            & 135.7           & 16.6     & 30.4     & 97.2   & 22.9   & 6.9      & 87.6   & 343.8  \\
                         & Imaging                   & 530.1            & 89.7            & -         & -         & -       & -       & -         & -       & -      \\
\bottomrule
\end{tabular}
}
\end{table*}

Table \ref{tab:time} shows the total runtime for each method in minutes. ITF-TAD's runtime is comparable to that of deep models. Although ITF-TAD avoids training neural networks, it requires considerable time for generating scalograms, mapping them into images, and selecting the coreset for PatchCore, which is used in image anomaly detection. Unlike deep models that require extensive time for hyperparameter optimization, ITF-TAD is practical as it does not require hyperparameter tuning. For SMD, which contains a large number of data points in each subdataset, the imaging computation time for ITF-TAD-PCA is particularly high, making ITF-TAD-RM a more practical choice.

is superior in terms of practicality. The computation times for GPT2-FPT and TimesNet are highly dependent on the stride of the time window used for anomaly detection. For SMD, the stride is set to 1, whereas for other datasets, it is set to 100. This results in much smaller runtime for SMD.

\section{Random Mapping}
\label{apx:rand}

\subsection{Random Matrix Generation}
\begin{figure}[t]
  \centering
  \includegraphics[width=0.9\linewidth]{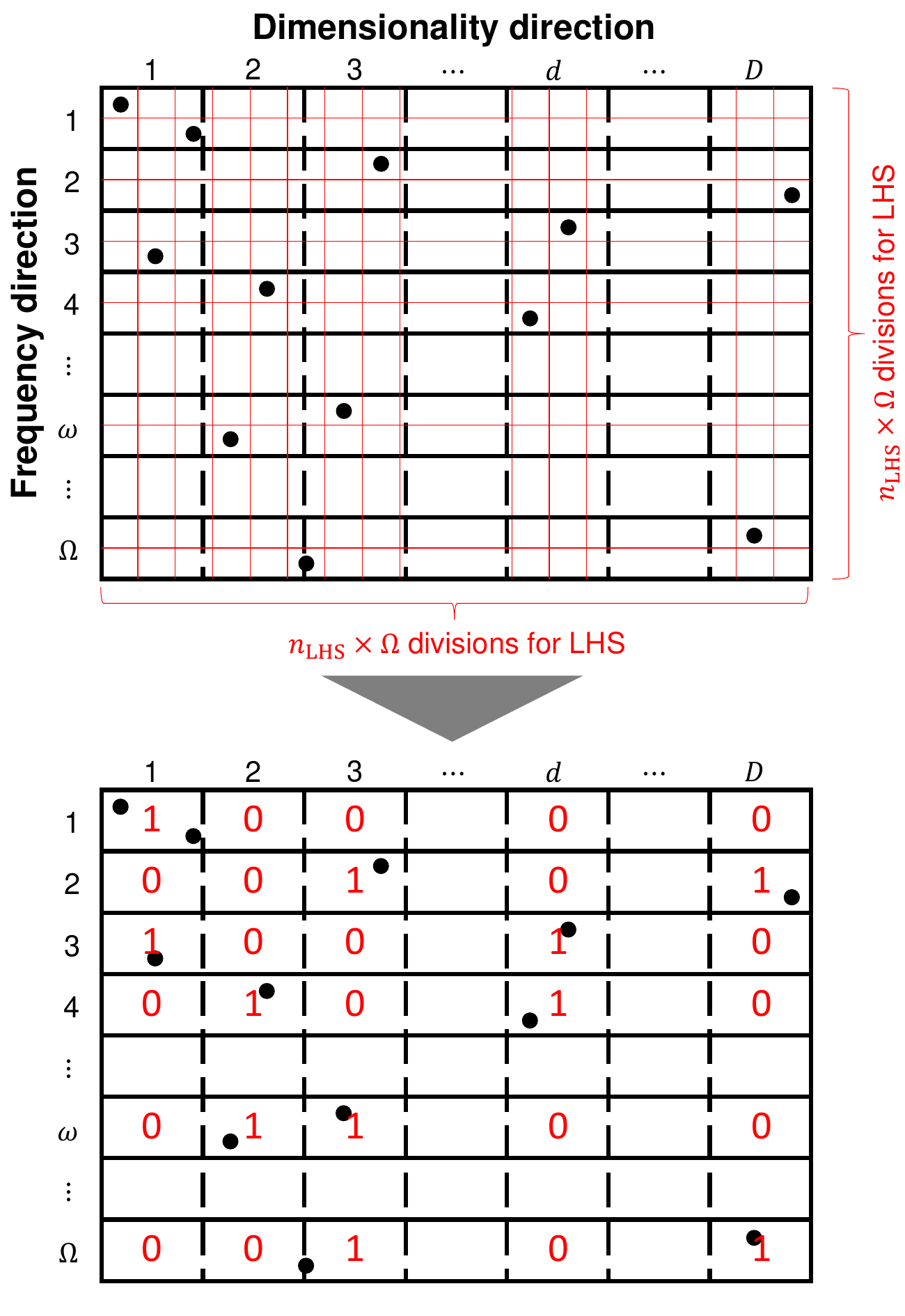}
  \caption{Random matrix generation using Latin hyper cube sampling}
  \label{fig:rand-matrix}
\end{figure}

The random matrix consisting of binary elements used in the random mapping is generated using Latin Hypercube Sampling (LHS). Figure \ref{fig:rand-matrix} illustrates the process where two elements that take the value of 1 are determined for each $\omega$ through LHS. When selecting dimensions less than or equal to $n_{\text{LHS}} = \max(2, \lfloor \nicefrac{D}{p} \rfloor)$ ($n_{\text{LHS}} = 2$ in Fig. \ref{fig:rand-matrix}) from each frequency, the first step is to generate $n_{\text{LHS}} \times \Omega$ sample points using LHS within the two-dimensional space $[0,\Omega] \times [0,D]$. Next, in this two-dimensional space, if a sample point exists within each region divided by $\Omega$ and $D$, 1 is assigned; otherwise, 0 is assigned. This process yields the random matrix $\Gamma = (\gamma_{i,d,\omega})$ used for scalograms with the $i$-th mother wavelet $\phi_i$. The condition that the number of elements taking the value of 1 for each frequency $\omega$ is less than or equal to $n_{\text{LHS}}$ arises because, as seen in Figure \ref{fig:rand-matrix} at ($\omega$, $d$) = (1, 1), when multiple sample points are generated within the same region, the count may fall below $n_{\text{LHS}}$. In this way, each dimension can be mapped equally to the frequencies.

\subsection{Effects of Random Matrix Sparsity on Performance}
The random matrix $\Gamma$ in Eq. \ref{eq:aggregate} contains $\max(2, \lfloor \nicefrac{D}{p} \rfloor)$ elements with 1 in the $\omega$-th column, where $p=5$ is used in the main results. We investigated the effect of $p$, which controls the sparsity of $\Gamma$, across the PSM, SMAP, MSL, and SMD datasets as shown in Table \ref{tab:rand}. Since $\Gamma$ is not used in univariate time series, the UCR dataset is excluded from this investigation. When $p=\infty$, two elements in each column are set to one, regardless of the number of dimensions $D$. Each experiment was repeated five times with different random seeds, using the average results for comparison.
Although the impact of $p$ is limited, there was a noticeable performance drop for the MSL dataset when $p=\infty$. This likely occurred because MSL has the highest number of dimensions (55), and selecting only 2 dimensions resulted in insufficient information being embedded in the images.

\begin{table}[t]
\caption{Effects of random matrix parameter $p$ on four multivariate benchmark datasets. The best results appear in bold.}
\label{tab:rand}
\centering
\scalebox{0.8}{
\begin{tabular}{cc|cccc}
\toprule
\multirow{2}{*}{Dataset} & \multirow{2}{*}{Metric} & \multicolumn{4}{c}{ITF-TAD-RM}                                    \\
                         &                         & $p=2$            & $p=5$            & $p=10$           & $p=\infty$          \\
\midrule
\multirow{2}{*}{PSM}     & F1*-SP                  & 0.882          & \textbf{0.889} & 0.869          & 0.869          \\
                         & AUCPR-SP                & 0.890           & 0.895          & \textbf{0.907} & \textbf{0.907} \\
\midrule
\multirow{2}{*}{SMAP}    & F1*-SP                  & 0.715          & \textbf{0.730}  & 0.720           & 0.720           \\
                         & AUCPR-SP                & 0.781          & \textbf{0.798} & 0.794          & 0.794          \\
\midrule
\multirow{2}{*}{MSL}     & F1*-SP                  & \textbf{0.725} & 0.724          & 0.696          & 0.652          \\
                         & AUCPR-SP                & 0.744          & \textbf{0.756} & 0.730           & 0.705          \\
\midrule
\multirow{2}{*}{SMD}     & F1*-SP                  & 0.643          & 0.660           & \textbf{0.671} & 0.667          \\
                         & AUCPR-SP                & 0.599          & 0.620           & 0.636          & \textbf{0.642} \\
\bottomrule
\end{tabular}
}
\end{table}

\section{Effects of Window Size in Score Partitioning}
\label{apx:sp}
Given that SP metrics depend on the window size $n_{sp}$, we conducted a parametric study on this variable. Table \ref{tab:sp} shows metrics evaluated at different window sizes $n_{sp}=50, 100, 256, \lfloor \nicefrac{T}{100} \rfloor, \lfloor \nicefrac{T}{1000} \rfloor$ where $T$ is the number of test data points in each subdataset. 
In most cases, the ranking among models remained consistent. However, for the SMAP and MSL datasets, GDN surpasses ITF-TAD only when $n_{sp}=\lfloor \nicefrac{D}{1000} \rfloor$. For SMAP and MSL, where $D \approx 5000$, this results in $n_{sp} \approx 5$. Such a small window size nearly eliminates the benefit of SP, leaving only the effects of PA.

\section{Comparison Using Other Evaluation Metrics}
\label{apx:metrics}
To further evaluate our models, we used three variants of key metrics: the best F1 score (F1*), AUCPR, and the area under the receiver operating characteristics curve (AUROC). These variations include metrics without any adjustments, with a PA adjustment, and with a SP adjustment, denoted by suffixes such as F1*-PA and F1*-SP. All metrics were calculated for each problem in the dataset, and their average values were used for comparison. 
For the UCR dataset, we focused on the number of correct answers. Since participants in the KDD Cup tuned their models specifically for the UCR dataset, and some models assumed only one anomaly per problem, a direct comparison of correct answers with KDD Cup is not feasible. 

Table \ref{tab:full_results} presents a performance comparison of each model using these evaluation metrics. In UCR, SMAP, and MSL, ITF-TAD demonstrated excellent results in metrics without any adjustment. The PSM dataset, which lacks subdatasets, contains a mix of long and short anomalies, as shown in Fig. \ref{fig:psm_full}. GDN and TranAD achieved the highest two values in unadjusted metrics for PSM because their anomaly scores broadly increased during long anomalies. However, as the figure indicates, ITF-TAD excels in detecting short anomalies, making SP-adjusted metrics more appropriate for comparison. In SMD, since the anomaly labels are short, the trends in SP-adjusted and PA-adjusted metrics are similar.

\section{Additional Visualization}
\label{apx:vis}
\begin{figure*}[tb]
  \centering
  \includegraphics[width=0.75\linewidth]{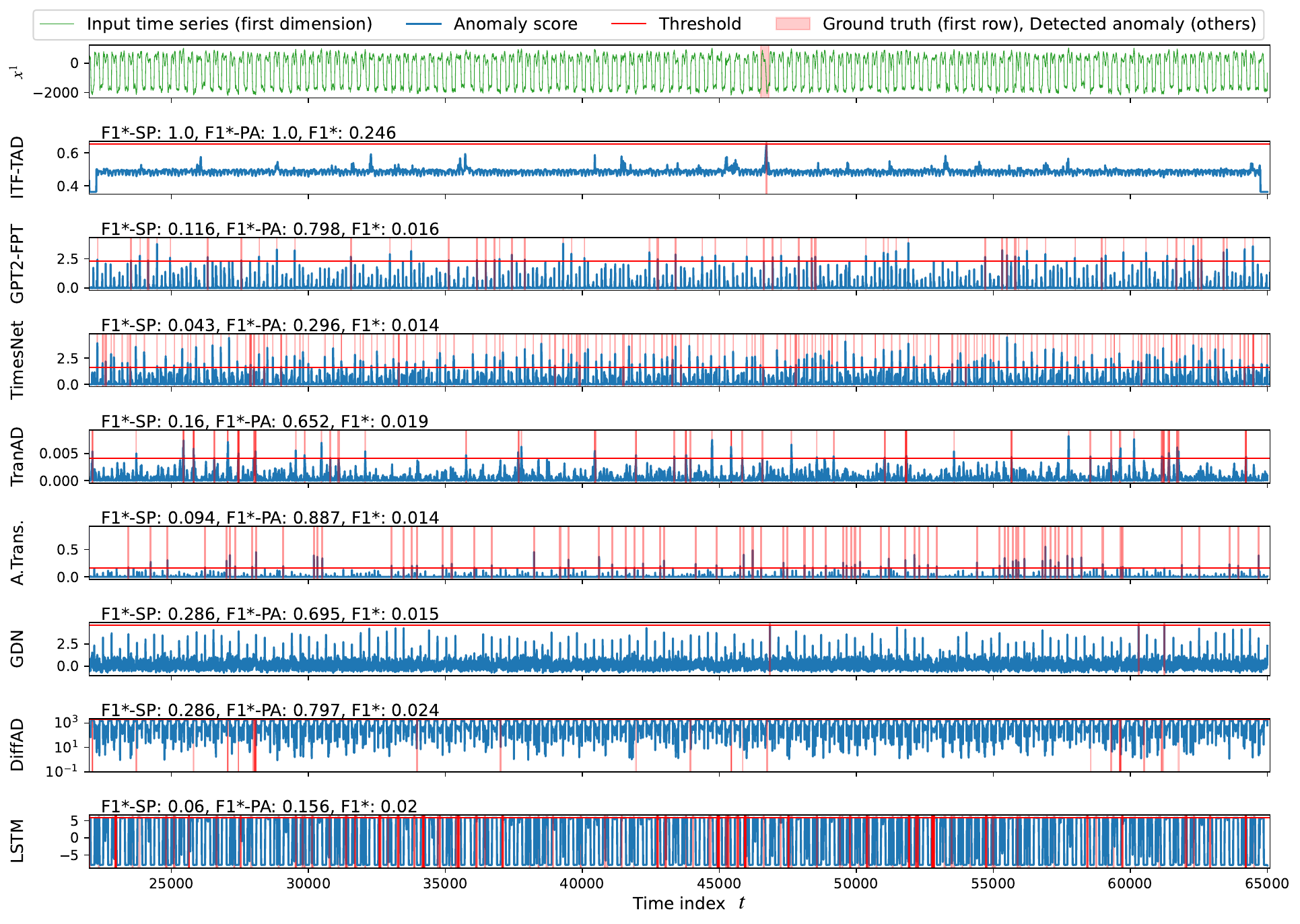}
  \caption{Anomaly scores from all models for the test section of UCR-060}
  \label{fig:ucr_full}
\end{figure*}

\begin{figure*}[tb]
  \centering
  \includegraphics[width=0.75\linewidth]{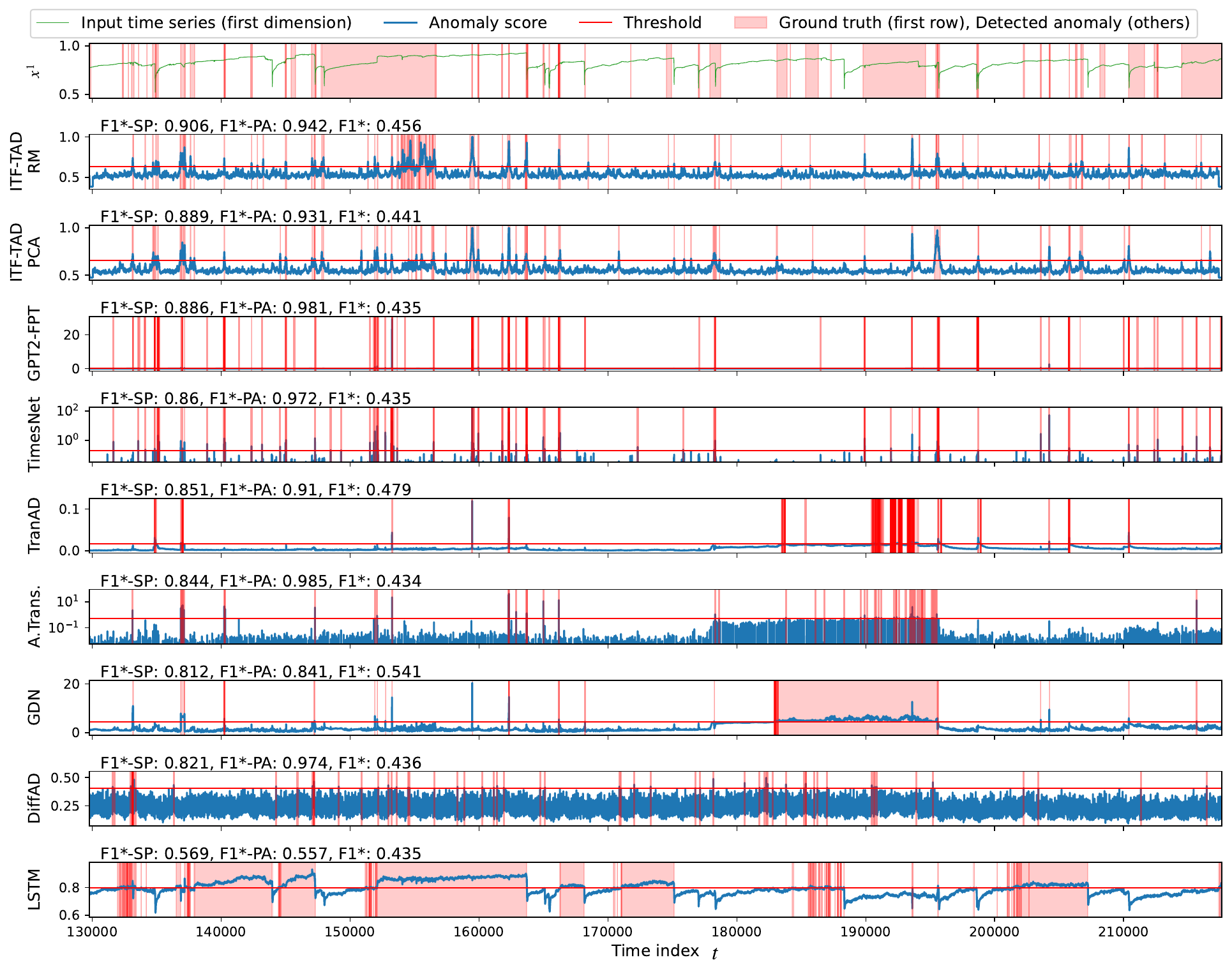}
  \caption{Anomaly scores from all models for the test section of PSM}
  \label{fig:psm_full}
\end{figure*}

\begin{figure*}[tb]
  \centering
  \includegraphics[width=0.75\linewidth]{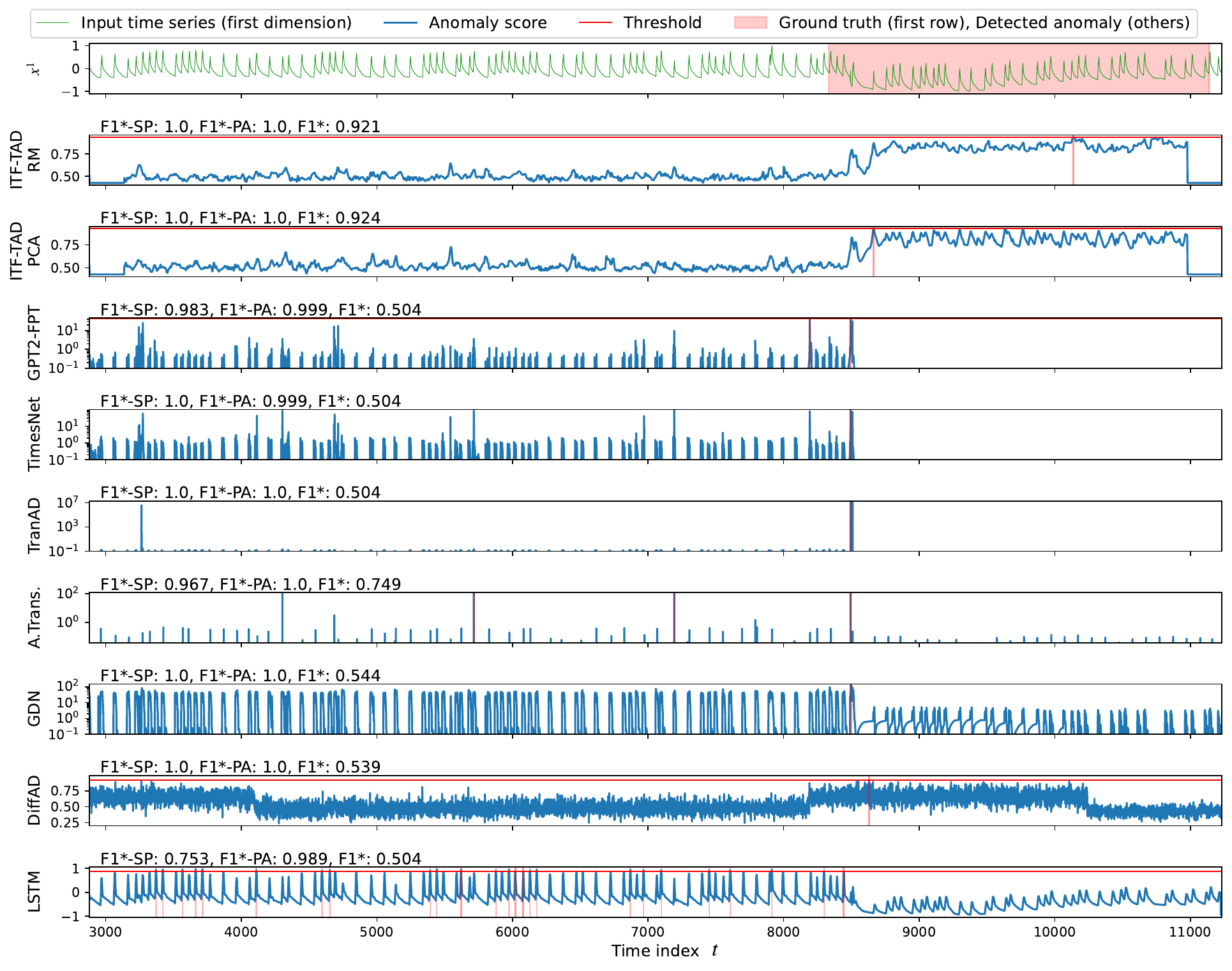}
  \caption{Anomaly scores from all models for the test section of SMAP-E-4}
  \label{fig:smap_full}
\end{figure*}

\begin{figure*}[tb]
  \centering
  \includegraphics[width=0.75\linewidth]{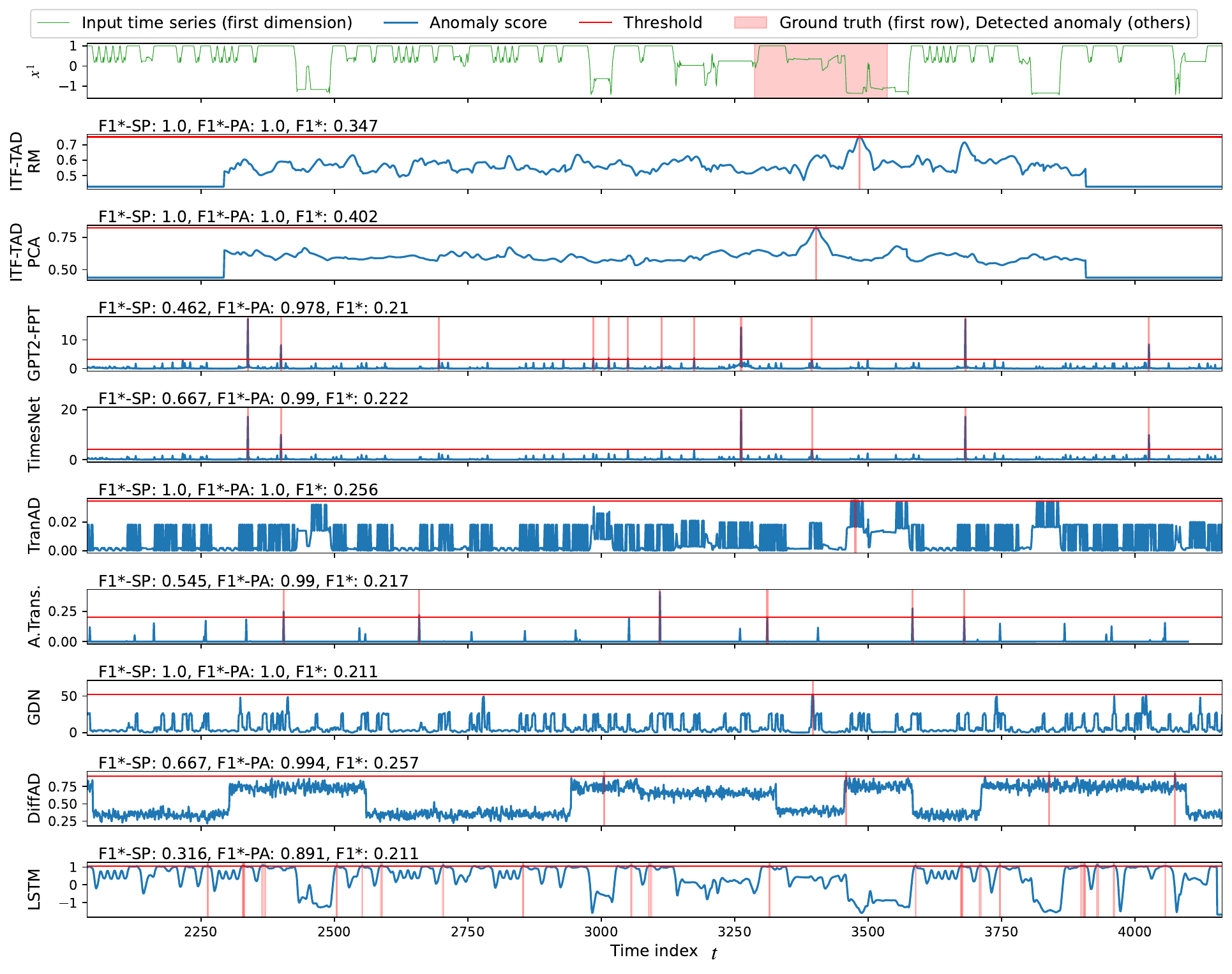}
  \caption{Anomaly scores from all models for the test section of MSL-M-3}
  \label{fig:msl_full}
\end{figure*}

\begin{figure*}[tb]
  \centering
  \includegraphics[width=0.75\linewidth]{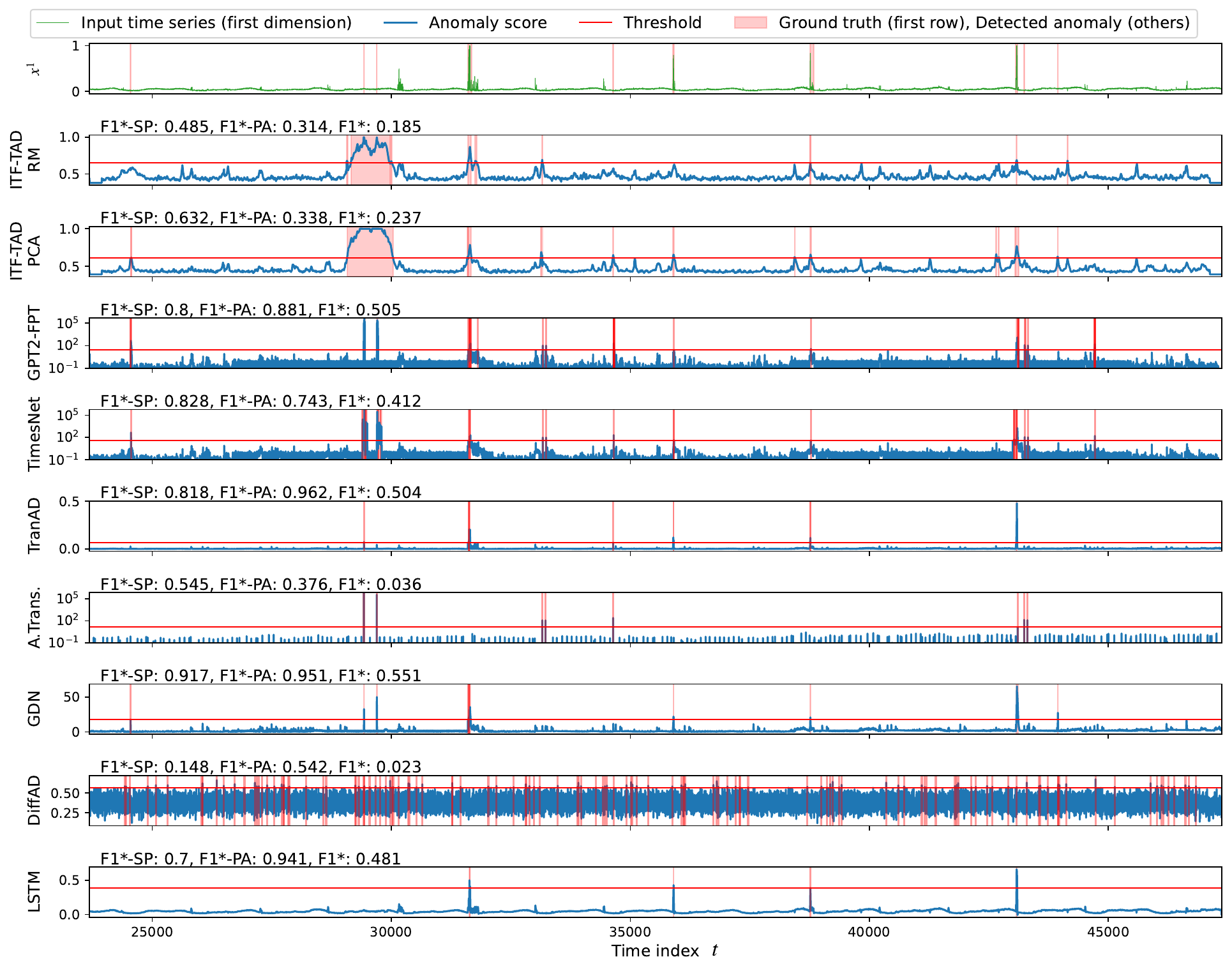}
  \caption{Anomaly scores from all models for the test section of SMD-2-3}
  \label{fig:smd_full}
\end{figure*}

To effectively compare the models, a representative subdataset from each dataset was chosen, and the anomaly scores were visualized in Figs. \ref{fig:ucr_full}-\ref{fig:smd_full}. Figs. \ref{fig:psm_full}-\ref{fig:msl_full} clearly show that applying PA is essential for a valid comparison in datasets with long anomaly labels. However, PA has inherent flaws. Therefore, using SP adjustments, which address these flaws, provides the most reliable comparison.

\begin{table*}[t]
\caption{Performance comparison with different window sizes for score partitioning. The best results are underlined, while the top two results appear in bold.}
\label{tab:sp}
\centering
\scalebox{0.65}{
\begin{tabular}{cc|cc|cc|cc|cc|cc}
\toprule
\multirow{2}{*}{$n_{sp}$}  & \multirow{2}{*}{dataset} & UCR                                   & UCR                                   & PSM                  & PSM                  & SMAP                 & SMAP                 & MSL                  & MSL                  & SMD                  & SMD                  \\
                       &                          & F1*-SP                                 & AUCPR-SP                              & F1*-SP                & AUCPR-SP             & F1*-SP                & AUCPR-SP             & F1*-SP                & AUCPR-SP             & F1*-SP                & AUCPR-SP             \\
\midrule
\multirow{9}{*}{256}   & ITF-TAD-PCA              & \multirow{2}{*}{\underline{ \textbf{0.732}}} & \multirow{2}{*}{\underline{ \textbf{0.839}}} & 0.854                & 0.897                & 0.749                & 0.779                & \underline{ \textbf{0.812}} & \underline{ \textbf{0.821}} & 0.703                & 0.677                \\
                       & ITF-TAD-RM               &                                       &                                       & \textbf{0.858}       & 0.875                & \textbf{0.756}       & \underline{ \textbf{0.795}} & 0.76                 & \textbf{0.786}       & 0.704                & 0.684                \\
                       & GPT2-FPT                 & 0.256                                 & 0.522                                 & \underline{ \textbf{0.859}} & \underline{ \textbf{0.929}} & 0.673                & 0.695                & 0.66                 & 0.688                & \textbf{0.838}       & \underline{ \textbf{0.829}} \\
                       & TimesNet                 & 0.262                                 & 0.535                                 & 0.827                & \textbf{0.912}       & 0.637                & 0.700                & \textbf{0.763}       & 0.765                & \underline{ \textbf{0.842}} & \textbf{0.822}       \\
                       & DiffAD                   & 0.227                                 & 0.507                                 & 0.737                & 0.771                & 0.564                & 0.641                & 0.622                & 0.667                & 0.393                & 0.282                \\
                       & TranAD                   & 0.319                                 & 0.564                                 & 0.800                & 0.889                & 0.749                & \textbf{0.793}       & 0.745                & 0.774                & 0.807                & 0.798                \\
                       & A.Trans.                 & 0.242                                 & 0.530                                 & 0.790                & 0.893                & 0.580                & 0.664                & 0.629                & 0.650                & 0.615                & 0.578                \\
                       & GDN                      & \textbf{0.514}                        & \textbf{0.697}                        & 0.809                & 0.877                & \underline{ \textbf{0.778}} & 0.788         & 0.741                & 0.768                & 0.821                & 0.802                \\
                       & LSTM                     & 0.225                                 & 0.513                                 & 0.638                & 0.510                & 0.659                & 0.702                & 0.756                & 0.771                & 0.555                & 0.523                \\
\midrule
\multirow{9}{*}{100}   & ITF-TAD-PCA              & \multirow{2}{*}{\underline{ \textbf{0.724}}} & \multirow{2}{*}{\underline{ \textbf{0.819}}} & \underline{ \textbf{0.889}} & 0.915                & 0.72                 & 0.797                & \underline{ \textbf{0.727}} & 0.737                & 0.645                & 0.585                \\
                       & ITF-TAD-RM               &                                       &                                       & \underline{ \textbf{0.889}} & 0.895                & 0.73                 & \textbf{0.798}       & 0.724                & \textbf{0.756}       & 0.660                 & 0.620                 \\
                       & GPT2-FPT                 & 0.231                                 & 0.504                                 & 0.886                & \underline{ \textbf{0.949}} & 0.651                & 0.719                & 0.606                & 0.637                & \textbf{0.844}       & \underline{ \textbf{0.833}} \\
                       & TimesNet                 & 0.244                                 & 0.521                                 & 0.860                & \textbf{0.934}       & 0.628                & 0.715                & 0.659                & 0.672                & \underline{ \textbf{0.850}}  & \textbf{0.829}       \\
                       & DiffAD                   & 0.199                                 & 0.487                                 & 0.821                & 0.842                & 0.518                & 0.642                & 0.598                & 0.626                & 0.330                 & 0.266                \\
                       & TranAD                   & 0.306                                 & 0.560                                 & 0.851                & 0.901                & \textbf{0.731}       & \underline{ \textbf{0.803}} & 0.713         & 0.751                & 0.800                 & 0.790                 \\
                       & A.Trans.                 & 0.237                                 & 0.526                                 & 0.844                & 0.913                & 0.577                & 0.681                & 0.575                & 0.616                & 0.602                 & 0.542                \\
                       & GDN                      & \textbf{0.512}                        & \textbf{0.671}                        & 0.812                & 0.883                & \underline{ \textbf{0.750}}  & \textbf{0.798}       & \underline{ \textbf{0.727}} & \underline{ \textbf{0.768}} & 0.808                & 0.779                \\
                       & LSTM                     & 0.197                                 & 0.513                                 & 0.569                & 0.435                & 0.620                & 0.685                & 0.719                & 0.727                & 0.571                & 0.514                \\
\midrule
\multirow{9}{*}{50}    & ITF-TAD-PCA              & \multirow{2}{*}{\underline{ \textbf{0.727}}} & \multirow{2}{*}{\underline{ \textbf{0.824}}} & \underline{ \textbf{0.907}} & 0.914                & 0.732                & 0.806                & 0.686                & 0.735                & 0.625                & 0.55                 \\
                       & ITF-TAD-RM               &                                       &                                       & \textbf{0.906}       & 0.902                & 0.738                & 0.804                & 0.704                & 0.743                & 0.661                & 0.604                \\
                       & GPT2-FPT                 & 0.256                                 & 0.529                                 & 0.901                & \underline{ \textbf{0.965}} & 0.668                & 0.727                & 0.611                & 0.661                & \underline{ \textbf{0.865}} & \underline{ \textbf{0.855}} \\
                       & TimesNet                 & 0.271                                 & 0.548                                 & 0.879                & \textbf{0.953}       & 0.642                & 0.724                & 0.643                & 0.673                & \textbf{0.850}        & \textbf{0.821}       \\
                       & DiffAD                   & 0.214                                 & 0.497                                 & 0.862                & 0.885                & 0.545                & 0.660                & 0.603                & 0.639                & 0.363                & 0.286                \\
                       & TranAD                   & 0.328                                 & 0.572                                 & 0.872                & 0.912                & \textbf{0.741}       & \underline{ \textbf{0.823}} & \textbf{0.717}       & \textbf{0.768}       & 0.815                & 0.806                \\
                       & A.Trans.                 & 0.269                                 & 0.550                                 & 0.867                & 0.936                & 0.588                & 0.712                & 0.553                & 0.612                & 0.645                & 0.580                 \\
                       & GDN                      & \textbf{0.530}                        & \textbf{0.689}                        & 0.817                & 0.889                & \underline{ \textbf{0.762}} & \textbf{0.810} & \underline{ \textbf{0.732}} & \underline{ \textbf{0.778}} & 0.821                & 0.788                \\
                       & LSTM                     & 0.196                                 & 0.520                                 & 0.558                & 0.422                & 0.639                & 0.707                & 0.716                & 0.739                & 0.605                & 0.536                \\
\midrule
\multirow{9}{*}{$\lfloor T/100 \rfloor$}  & ITF-TAD-PCA              & \multirow{2}{*}{\underline{ \textbf{0.725}}} & \multirow{2}{*}{\underline{ \textbf{0.835}}} & 0.871                & \textbf{0.943}       & 0.723                & \textbf{0.800}         & 0.700                  & 0.714                & 0.684                & 0.638                \\
                       & ITF-TAD-RM               &                                       &                                       & 0.848                & 0.925                & \textbf{0.731}       & 0.792                & 0.706                & 0.742                & 0.668                & 0.646                \\
                       & GPT2-FPT                 & 0.237                                 & 0.513                                 & \underline{ \textbf{0.893}} & \underline{ \textbf{0.958}} & 0.656                & 0.718                & 0.665                & 0.687                & \underline{ \textbf{0.819}} & \underline{ \textbf{0.809}} \\
                       & TimesNet                 & 0.248                                 & 0.534                                 & 0.850                & 0.936                & 0.629                & 0.715                & 0.700                & 0.696                & \textbf{0.814}       & \textbf{0.792}       \\
                       & DiffAD                   & 0.205                                 & 0.501                                 & 0.790                & 0.769                & 0.532                & 0.637                & 0.632                & 0.667                & 0.357                & 0.293                \\
                       & TranAD                   & 0.298                                 & 0.562                                 & 0.821                & 0.919                & \textbf{0.731}       & 0.798                & \underline{ \textbf{0.769}} & \textbf{0.782}       & 0.792                & 0.789                \\
                       & A.Trans.                 & 0.228                                 & 0.508                                 & 0.839                & 0.917                & 0.582                & 0.682                & 0.593                & 0.632                & 0.582                & 0.546                \\
                       & GDN                      & \textbf{0.503}                        & \textbf{0.687}                        & \textbf{0.878}       & 0.917                & \underline{ \textbf{0.752}} & \underline{ \textbf{0.802}} & \textbf{0.762}       & 0.780                & 0.813                & 0.791                \\
                       & LSTM                     & 0.204                                 & 0.504                                 & 0.800                & 0.744                & 0.631                & 0.695                & 0.747                & \underline{ \textbf{0.790}}  & 0.539                & 0.513                \\
\midrule
\multirow{9}{*}{$\lfloor T/1000 \rfloor$} & ITF-TAD-PCA              & \multirow{2}{*}{\underline{ \textbf{0.728}}} & \multirow{2}{*}{\underline{ \textbf{0.812}}} & \textbf{0.891}       & \underline{ \textbf{0.951}} & 0.764                & 0.834                & 0.738                & 0.781                & 0.632                & 0.548                \\
                       & ITF-TAD-RM               &                                       &                                       & \underline{ \textbf{0.893}} & 0.94                 & 0.763                & 0.820                 & 0.743                & 0.781                & 0.680                 & 0.610                 \\
                       & GPT2-FPT                 & 0.321                                 & 0.583                                 & 0.883                & \textbf{0.950}        & 0.748                & 0.787                & 0.847                & 0.823                & \underline{ \textbf{0.872}} & \underline{ \textbf{0.853}} \\
                       & TimesNet                 & 0.310                                 & 0.562                                 & 0.860                & 0.938                & 0.736                & 0.783                & 0.856                & 0.827                & \textbf{0.854}       & 0.814                \\
                       & DiffAD                   & 0.246                                 & 0.526                                 & 0.822                & 0.850                & 0.677                & 0.754                & 0.805                & 0.845                & 0.431                & 0.344                \\
                       & TranAD                   & 0.358                                 & 0.598                                 & 0.854                & 0.901                & \textbf{0.813}       & \textbf{0.865}       & \textbf{0.873}       & \underline{ \textbf{0.894}} & 0.835                & \textbf{0.827}       \\
                       & A.Trans.                 & 0.332                                 & 0.602                                 & 0.846                & 0.910                & 0.717                & 0.787                & 0.776                & 0.831                & 0.692                & 0.623                \\
                       & GDN                      & \textbf{0.562}                        & \textbf{0.714}                        & 0.812                & 0.883                & \underline{ \textbf{0.828}} & \underline{ \textbf{0.869}} & \underline{ \textbf{0.881}} & \textbf{0.893}       & 0.839                & 0.808                \\
                       & LSTM                     & 0.233                                 & 0.540                                 & 0.568                & 0.431                & 0.724                & 0.783                & 0.820                 & 0.848                & 0.642                & 0.566         \\
\bottomrule
\end{tabular}
}
\end{table*}

\begin{table*}[t]
\caption{Performance comparison using 10 metrics among ITF-TAD and baselines on five benchmark datasets. The best results are underlined, while the top two results appear in bold.}
\label{tab:full_results}
\centering
\scalebox{0.75}{
\begin{tabular}{cc|ccccccccc}
\toprule
\multirow{2}{*}{Dataset} & \multirow{2}{*}{Metric} & \multicolumn{2}{c}{ITF-TAD (Ours)}          & GPT2-FPT             & TimesNet             & DiffAD         & TranAD               & A.Trans.             & GDN                  & LSTM   \\
                         &                         & PCA                  & RM                   & (2023)               & (2023)               & (2023)         & (2022)               & (2022)               & (2021)               & (2018) \\
\midrule
\multirow{10}{*}{UCR}    & F1*-SP                  & \multicolumn{2}{c}{\underline{ \textbf{0.724}}}    & 0.231                & 0.244                & 0.199          & 0.306                & 0.237                & \textbf{0.512}       & 0.197  \\
                         & AUCPR-SP                & \multicolumn{2}{c}{\underline{ \textbf{0.819}}}    & 0.504                & 0.521                & 0.487          & 0.560                & 0.526                & \textbf{0.671}       & 0.513  \\
                         & Correct                 & \multicolumn{2}{c}{\underline{ \textbf{0.927}}}    & 0.672                & 0.655                & 0.658          & 0.682                & 0.684                & \textbf{0.808}       & 0.613  \\
                         & F1*-PA                  & \multicolumn{2}{c}{\underline{ \textbf{0.744}}}    & 0.531                & 0.496                & 0.402          & 0.497                & 0.644                & \textbf{0.684}       & 0.342  \\
                         & AUCPR-PA                & \multicolumn{2}{c}{\underline{ \textbf{0.850}}}    & 0.728                & 0.712                & 0.668          & 0.719                & 0.788                & \textbf{0.815}       & 0.640  \\
                         & AUROC-PA                & \multicolumn{2}{c}{\underline{ \textbf{0.959}}}    & 0.928                & 0.923                & 0.898          & 0.911                & 0.918                & \textbf{0.947}       & 0.828  \\
                         & F1*                     & \multicolumn{2}{c}{\underline{ \textbf{0.461}}}    & 0.037                & 0.044                & 0.043          & 0.052                & 0.025                & \textbf{0.141}       & 0.037  \\
                         & AUCPR                   & \multicolumn{2}{c}{\underline{ \textbf{0.424}}}    & 0.045                & 0.051                & 0.049          & 0.053                & 0.092                & \textbf{0.118}       & 0.088  \\
                         & AUROC                   & \multicolumn{2}{c}{\underline{ \textbf{0.900}}}    & 0.529                & 0.539                & 0.562          & 0.529                & 0.502                & \textbf{0.593}       & 0.491  \\
                         & Correct                 & \multicolumn{2}{c}{\underline{ \textbf{150}}}      & 22                   & 33                   & 23             & 37                   & 17                   & \textbf{99}          & 20     \\
\midrule
\multirow{9}{*}{PSM}     & F1*-SP                  & \underline{ \textbf{0.889}} & \underline{ \textbf{0.889}} & 0.886                & 0.860                & 0.821          & 0.851                & 0.844                & 0.812                & 0.569  \\
                         & AUCPR-SP                & 0.915                & 0.895                & \underline{ \textbf{0.949}} & \textbf{0.934}       & 0.842          & 0.901                & 0.913                & 0.883                & 0.435  \\
                         & AUROC-SP                & \textbf{0.961}       & 0.949                & \underline{ \textbf{0.963}} & 0.949                & 0.894          & 0.912                & 0.932                & 0.928                & 0.660  \\
                         & F1*-PA                  & 0.931                & 0.938                & \textbf{0.981}       & 0.972                & 0.974          & 0.910                & \underline{ \textbf{0.985}} & 0.841                & 0.557  \\
                         & AUCPR-PA                & 0.893                & 0.925                & \underline{ \textbf{0.995}} & \textbf{0.994}       & 0.992          & 0.937                & 0.993                & 0.912                & 0.443  \\
                         & AUROC-PA                & 0.987                & 0.988                & \underline{ \textbf{0.998}} & \underline{ \textbf{0.998}} & 0.996          & 0.955                & 0.992                & 0.960                & 0.727  \\
                         & F1*                     & 0.441                & 0.444                & 0.435                & 0.435                & 0.436          & \textbf{0.479}       & 0.434                & \underline{ \textbf{0.541}} & 0.435  \\
                         & AUCPR                   & 0.384                & 0.426                & 0.394                & 0.391                & 0.318          & \textbf{0.456}       & 0.287                & \underline{ \textbf{0.500}} & 0.234  \\
                         & AUROC                   & 0.603                & 0.620                & 0.595                & 0.593                & 0.553          & \textbf{0.640}       & 0.502                & \underline{ \textbf{0.721}} & 0.395  \\
\midrule
\multirow{9}{*}{SMAP}    & F1*-SP                  & 0.720                & 0.730                & 0.651                & 0.628                & 0.518          & \textbf{0.731}       & 0.577                & \underline{ \textbf{0.750}}  & 0.620   \\
                         & AUCPR-SP                & 0.797                & \textbf{0.798}       & 0.719                & 0.715                & 0.642          & \underline{ \textbf{0.803}} & 0.681         & \textbf{0.798}       & 0.685  \\
                         & AUROC-SP                & \underline{ \textbf{0.873}} & 0.865         & 0.840                & 0.830                & 0.777          & 0.863                & 0.771                & \textbf{0.869}       & 0.758  \\
                         & F1*-PA                  & 0.777                & 0.773                & 0.836                & 0.844                & 0.814          & \underline{ \textbf{0.883}} & 0.850         & \textbf{0.877}       & 0.769  \\
                         & AUCPR-PA                & 0.841                & 0.839                & 0.857                & 0.856                & 0.877          & \underline{ \textbf{0.909}} & 0.898         & \textbf{0.902}       & 0.836  \\
                         & AUROC-PA                & 0.949                & 0.933                & 0.973                & 0.974                & \textbf{0.985} & 0.979                & 0.981                & \underline{ \textbf{0.988}} & 0.941  \\
                         & F1*                     & \underline{ \textbf{0.478}} & \textbf{0.473} & 0.338               & 0.343                & 0.230          & 0.391                & 0.247                & 0.342                & 0.374  \\
                         & AUCPR                   & \underline{ \textbf{0.451}} & \textbf{0.436} & 0.225               & 0.226                & 0.148          & 0.299                & 0.177                & 0.223                & 0.294  \\
                         & AUROC                   & \underline{ \textbf{0.761}} & \textbf{0.753} & 0.548               & 0.608                & 0.585          & 0.641                & 0.550                & 0.603                & 0.572  \\
\midrule
\multirow{9}{*}{MSL}     & F1*-SP                  & \underline{ \textbf{0.727}} & 0.724         & 0.606                & 0.659                & 0.598          & 0.713                & 0.575                & \underline{ \textbf{0.727}} & 0.719  \\
                         & AUCPR-SP                & 0.737                & \textbf{0.756}       & 0.637                & 0.672                & 0.626          & 0.751                & 0.616                & \underline{ \textbf{0.768}} & 0.727  \\
                         & AUROC-SP                & \underline{ \textbf{0.844}} & 0.822         & 0.775                & 0.787                & 0.702          & \textbf{0.831}       & 0.744                & 0.828                & 0.788  \\
                         & F1*-PA                  & 0.743                & 0.749                & 0.883                & 0.886                & 0.848          & \textbf{0.896}       & 0.821                & \underline{ \textbf{0.900}}   & 0.835  \\
                         & AUCPR-PA                & 0.785                & 0.786                & 0.872                & 0.872                & 0.879          & \underline{ \textbf{0.913}} & 0.875                & \textbf{0.910}        & 0.861  \\
                         & AUROC-PA                & 0.882                & 0.882                & 0.968                & 0.970                & 0.974          & \underline{ \textbf{0.984}} & 0.965                & \textbf{0.983}       & 0.922  \\
                         & F1*                     & \underline{ \textbf{0.411}} & 0.362          & 0.320               & 0.377                & 0.316          & \textbf{0.406}       & 0.203                & 0.376                & 0.367  \\
                         & AUCPR                   & \underline{ \textbf{0.334}} & \textbf{0.291} & 0.230               & 0.278                & 0.214          & 0.281                & 0.123                & 0.272                & 0.272  \\
                         & AUROC                   & \underline{ \textbf{0.721}} & \textbf{0.688} & 0.631               & 0.676                & 0.586          & 0.629                & 0.434                & 0.641                & 0.567  \\
\midrule
\multirow{9}{*}{SMD}     & F1*-SP                  & 0.645                & 0.660                & \textbf{0.844}       & \underline{ \textbf{0.850}}  & 0.330           & 0.800                  & 0.602                & 0.808                & 0.571  \\
                         & AUCPR-SP                & 0.585                & 0.620                & \underline{ \textbf{0.833}} & \textbf{0.829}       & 0.266          & 0.790                 & 0.542                & 0.779                & 0.514  \\
                         & AUROC-SP                & 0.919                & 0.907                & \underline{ \textbf{0.969}} & \underline{ \textbf{0.969}} & 0.672          & 0.950                 & 0.799                & 0.959                & 0.770   \\
                         & F1*-PA                  & 0.663                & 0.723                & \underline{ \textbf{0.930}}  & 0.908                & 0.793          & \textbf{0.910}        & 0.860                 & 0.901                & 0.716  \\
                         & AUCPR-PA                & 0.558                & 0.648                & \underline{ \textbf{0.911}} & 0.881                & 0.751          & \textbf{0.898}       & 0.817                & 0.869                & 0.658  \\
                         & AUROC-PA                & 0.959                & 0.969                & \underline{ \textbf{0.997}} & \textbf{0.996}       & 0.985          & 0.991                & 0.950                 & 0.990                 & 0.894  \\
                         & F1*                     & 0.421                & 0.457                & 0.461                & \textbf{0.474}       & 0.090           & 0.450                 & 0.082                & \underline{ \textbf{0.504}} & 0.360   \\
                         & AUCPR                   & 0.333                & 0.406                & \textbf{0.412}       & 0.409                & 0.050           & 0.389                & 0.036                & \underline{ \textbf{0.444}} & 0.309  \\
                         & AUROC                   & \underline{ \textbf{0.892}} & \textbf{0.882}       & 0.849                & \textbf{0.882}       & 0.539          & 0.792                & 0.278                & 0.848                & 0.714 \\
\bottomrule
\end{tabular}
}
\end{table*}

\end{document}